\documentclass[journal]{IEEEtran}
\usepackage{graphicx}
\usepackage{booktabs}
\usepackage{multirow}
\usepackage{float}
\usepackage{mathtools}   %公式多行
\usepackage{pgfplots}
\usepackage{caption}
\usepackage{pgfplots}
\usepackage{subcaption}
\usepackage{array} 
\renewcommand\arraystretch{2}
\usepackage[T1]{fontenc}
\usepackage{amssymb}
\usepackage{algorithm}
\usepackage{algorithmic}
\newlength\myindent
\usepackage{etoolbox}
\AtBeginEnvironment{algorithmic}{\footnotesize}

\usepackage[driverfallback=dvips,colorlinks,bookmarksopen,bookmarksnumbered,citecolor=green,urlcolor=green]{hyperref}
\usepackage[square, comma, sort&compress, numbers]{natbib}   % 参考文献引用格式[4,9,10-12]
\usepackage{hyperref}
\hypersetup{
hypertex=true,
colorlinks=true,
linkcolor=blue,
anchorcolor=blue,
citecolor=blue
}

\ifCLASSINFOpdf
\else
\fi
\begin{document}
%
% paper title
% Titles are generally capitalized except for words such as a, an, and, as,
% at, but, by, for, in, nor, of, on, or, the, to and up, which are usually
% not capitalized unless they are the first or last word of the title.
% Linebreaks \\ can be used within to get better formatting as desired.
% Do not put math or special symbols in the title.
%\title{A Multi-Graph Convolutional Neural Networks based Method for Traffic Flow Prediction}
\title{FasterSTS: A Faster Spatio-Temporal Synchronous Graph Convolutional Networks for Traffic flow
Forecasting}
%
%
% author names and IEEE memberships
% note positions of commas and nonbreaking spaces ( ~ ) LaTeX will not break
% a structure at a ~ so this keeps an author's name from being broken across
% two lines.
% use \thanks{} to gain access to the first footnote area
% a separate \thanks must be used for each paragraph as LaTeX2e's \thanks
% was not built to handle multiple paragraphs
%

\author{Ben-Ao~Dai,~%~\IEEEmembership{Member,~IEEE,}
        Nengchao Lyu,~%~\IEEEmembership{Fellow,~OSA,}
        Yongchao Miao~
        %Huimin~Gao,~
        %Lingxi~Li,~
        %Weimin~Wu%,~\IEEEmembership{Senior Member,~IEEE}% <-this % stops a space % Hmgao@zjxu.edu.cn
\thanks{This work was supported by the National Natural Science Foundation of China under Grant 52072290 and the National Key Research and Development Program of China under Grant 2023YFB4302600.}% <-this % stops a space
%\thanks{Ben-Ao Dai and Mingjian Zhang are with the  School of Information Science and Engineering, Zhejiang Sci-Tech University, Xiasha Campus, Hangzhou, Zhejiang, 310018, China (email:mjzhang998@163.com).}
\thanks{Ben-Ao Dai is with Sanya science and education innovation park, Wuhan University of Technology,Sanya,Hainan,572000,China;and also with the Intelligent Transportation Systems Research Center, Wuhan University of Technology, Wuhan, Hubei, 430063, China; (email:Daiba1201@163.com).}
%\thanks{Bin Chen is with the School of Information Science and Engineering, Jiaxing University, Jiaxing, Zhejiang, 314001, China (email: Chenbin@zjxu.edu.cn).}
\thanks{Nengchao Lyu is a professor of Sanya science and education innovation park, Wuhan University of Technology,Sanya,Hainan,572000,China and The intelligent Transportation Systems Research Center, Wuhan University of Technology, Wuhan, Hubei, 430063, China. (e-mail: lnc@whut.edu.cn).}
\thanks{Yongchao Miao is with the School of Information Science and Engineering, Zhejiang Sci-Tech University, Hangzhou, Zhejiang, 310018, China; (email:Miaoyc0308@163.com).}
\thanks{Corresponding author:Nengchao Lyu(e-mail: lnc@whut.edu.cn).}
}
\maketitle
 % As a general rule, do not put math, special symbols or citations
% in the abstract or keywords.
\begin{abstract}
Accurate traffic flow prediction heavily relies on the spatio-temporal correlation of traffic flow data. Most current studies separately capture correlations in spatial and temporal dimensions, making it difficult to capture complex spatio-temporal heterogeneity, and often at the expense of increasing model complexity to improve prediction accuracy. Although there have been groundbreaking attempts in the field of spatio-temporal synchronous modeling, significant limitations remain in terms of performance and complexity control.This study proposes a quicker and more effective spatio-temporal synchronous traffic flow forecast model to address these issues.Firstly, the new graph computation method is reduced from the original graph computation method \begin{math}
\mathcal{O}(N2)
\end{math} to  \begin{math}
\mathcal{O}(KN)
\end{math} in terms of time complexity, thereby eliminating the traditional graph computation and lowering the computational complexity of both the graph convolution and the spatio-temporal synchronous graph convolution kernel.Additionally,Temporal correlation is captured during the graph convolution process by creating a novel spatio-temporal synchronous graph convolution kernel.Furthermore, the traffic network's static spatial correlation is initially acquired through data-driven methods in order to improve the model of spatial correlation. The dynamic spatial correlation is then captured by dynamically modifying the generated static adaptive graph using the dynamic spatio-temporal synchronous graph convolution kernel.The model maintains high prediction accuracy with low time complexity and resource consumption even when it does not use the essential elements of spatio-temporal modeling, such as the multi-head self-attention mechanism, recurrent neural networks (RNNs), and Time Convolutional Networks(TCNs). This facilitates the integration of reinforcement learning and other real-time demanding tasks with spatio-temporal modeling. The model tackles the challenge of simulating spatio-temporal correlations in mega-urban road networks and may be learned and implemented on devices with limited resources.
%

%Traffic flow forecasting is an important part of the intelligent transportation system, but due to traffic conditions, geographical location, time and other factors.Due to the influence of factors, it is highly nonlinear and complex, and it is difficult to achieve accurate prediction. Aiming at the problem of forecasting the traffic flow in and out of traffic stations, a Context Gating-Based Spatio-Temporal Multi-Graph Convolutional Network (CG-STMGCN) Model. Construct a neighbor graph according to the neighbor relationship and circulation flow relationship between sites.The circulation flow graph represents the proximity correlation and flow dependence between the site traffic, and the context-gated spatio-temporal convolution module is established on the two graphs.The spatio-temporal characteristics of site traffic are captured, and the outputs of the two graphs are fused using the Hadamard product as the final prediction result. On the real traffic station data set the experimental results show that the prediction accuracy of the CG-STMGCN model is better than that of similar prediction methods, and the stability is stronger.
\end{abstract}

% Note that keywords are not normally used for peerreview papers.
\begin{IEEEkeywords}
Spatio-temporal synchronous modeling; traffic flow prediction;Faster neural networks.
\end{IEEEkeywords}

\IEEEpeerreviewmaketitle

\section{Introduction}
% The very first letter is a 2 line initial drop letter followed
% by the rest of the first word in caps.
% 
% form to use if the first word consists of a single letter:
% \IEEEPARstart{A}{demo} file is ....
% 
% form to use if you need the single drop letter followed by
% normal text (unknown if ever used by the IEEE):
% \IEEEPARstart{A}{}demo file is ....
% 
% Some journals put the first two words in caps:
% \IEEEPARstart{T}{his demo} file is ....
% 
% Here we have the typical use of a "T" for an initial drop letter
% and "HIS" in caps to complete the first word.
%
%\IEEEPARstart{W}{ith} the rapid advancement of economic development and urbanization, the number of urban residents and car ownership has increased sharply, and urban roads are facing great challenges. In recent years, with the development of artificial intelligence, a new smart city blueprint is brewing. Intelligent Transportation System (ITS) \cite{1}, as a key element of urban smart management, can contribute new solutions to urban road traffic problems. Traffic forecasting is an important part of ITS, which can help traffic managers perceive and control road traffic conditions in advance, control and dispatch traffic flow, choose efficient travel routes for citizens, avoid possible road congestion in the future and save travel time . Although previous studies have achieved important results in traffic flow forecasting, it is difficult to accurately predict future road traffic conditions due to the variability and high nonlinearity of traffic flow.
%
\IEEEPARstart{T}{raffic} flow forecasting is an important component of the Intelligent Transportation System (ITS)\cite{0-1} and has become a hot research topic. Effective traffic forecasting methods are the basis for anticipating road traffic conditions and helping traffic managers to divert traffic congestion. They also help citizens to choose the most efficient travel routes, and can assist with road planning\cite{0-2} and autonomous driving\cite{1}.

Traffic flow forecasting is mainly divided into two categories: model-driven and data-driven. The key to model-driven methods lies in building an accurate forecasting model, but due to complexity and nonlinearity, it is often difficult to construct an effective model.
Data-driven methods\cite{4,5} are further classified into two approaches: statistical theory-based and machine learning-based. Statistical theory-based methods are simple and straightforward, yet they struggle to accurately predict highly nonlinear and complex traffic data. In contrast, machine learning-based methods\cite{10,11} boast a simpler model architecture and have significant advantages in handling large-scale and complex datasets compared to statistical theory-based methods.

As deep learning matures, it exhibits significant advantages in traffic flow forecasting. Recurrent Neural Networks (RNNs) excel at learning nonlinear traffic flow characteristics but suffer from gradient explosion or vanishing issues. To address this, Long Short-Term Memory (LSTM)\cite{12-0} and Gated Recurrent Unit (GRU)\cite{12-1} are used to extract temporal dependencies and handle long-term features. However, these methods neglect the spatial correlation of road networks. To improve prediction accuracy, researchers employ Convolutional Neural Networks (CNNs) to extract spatial correlations, such as the LSTM-CNN model integrated by Vijayalakshmi et al. in \cite{15}. Yet, CNNs have limited capability in modeling non-Euclidean dependencies.

In a recent study, Spatio-Temporal Graph Neural Networks (STGNNs) have become a hot topic in traffic flow forecasting research due to the outstanding ability to extract spatio-temporal correlations\cite{15-1,15-2}. STGNNs integrate the advantages of graph convolutional networks (GCN)\cite{15-3} and time series models \cite{15-6} in spatio-temporal modeling. GCN captures non-Euclidean spatial correlations, while temporal models capture temporal correlations. Many researchers have been working hard to design more powerful STGNNs to further improve the model's forecasting performance.In \cite{16}, a graph-based neural network model was presented to deal with the problem of classifying nodes in a graph.However, since the model is limited to undirected graphs, it is not suitable to treat topological information of traffic networks.
In \cite{19}, Yu et al. proposed a spatio-temporal graph convolutional network (STGCN) to complete the task of traffic flow forecasting. STGCN uses graph convolutional networks (GCNs) instead of CNN to capture comprehensive spatial correlation of different traffic nodes in traffic networks. 
In \cite{17}, the task of traffic flow forecasting has been modeled as a diffusion process on a directed graph with a diffusion convolutional recurrent neural network (DCRNN). In the deep learning framework of DCRNN, the spatial dependency of traffic flow was captured by a bidirectional diffusion convolution operation, while the temporal dependency of traffic flow was modeled by a encoder-decoder architecture with scheduled sampling. 
In \cite{18}, Wu et al. proposed Graph WaveNet, which preserves the hidden spatial dependency by constructing an adaptive graph. Simultaneously, it is possible to discover invisible graph structures from data without the guidance of any prior knowledge. Furthermore, some GCN and RNN-based hybrid forecasting models \cite{20} have been proposed to improve the forecasting accuracy. Since only some simple spatial dependencies such as whether the road network nodes were connected were taken into account, these hybrid forecasting models can not effectively extract enough spatial correlation of the road network.
In \cite{21-1}, Xu et al. reported  Spatio-Temporal Transformer Networks (STTNs) for traffic flow forecasting. In the STTNs, real-time traffic conditions and directions of traffic flow were modeled with a spatial transformer, while the long-range temporal dependencies across multiple time steps were caputred with a temporal
transformer.% However, the advantage of STTNs is mainly reflected in modeling long-distance features, and can not effectively capture some short-distance and local features.

%Although the above prediction methods model the spatial and temporal dependency in the traffic flow with various deep learning frameworks (e.g. RNN, CNN and GCN) and achieve some attractive results, most of them ignore the fact that spatio-temporal characteristics may dynamically change due to traffic congestion and the contextual information of time series. In addition, since traffic networks include rich spatio-temporal attributes, how to effectively model these spatio-temporal features of traffic flow from different perspectives is still one of the main challenges in the task of traffic flow forecasting.
%

As researchers improved and combined the widely used components in the time series forecasting models based on STGNNs\cite{17,18,19,20,21,21-1}  became increasingly complex but their performance improvements were limited, which led to a performance bottleneck in STGNNs to some extent \cite{15-2,33-1} , aiming to improve prediction accuracy by combining the strengths of different components in modeling spatio-temporal correlations. This neglects the heterogeneity of spatio-temporal data. In a recent study \cite{34,35}, a Spatio-Temporal Synchronous Graph Convolutional Networks (STSGCN) was proposed. STSGCN employs the fusion of temporal graphs and topological graphs to construct local spatio-temporal correlation graphs., which capture the spatio-temporal correlation in adjacent time steps of the road network. Since the constructed spatio-temporal graph is obtained by fusion, the time complexity increases exponentially compared to the topological adjacency graph. The local spatio-temporal graph is used to capture local spatio-temporal correlation, while long-term spatio-temporal correlation is realized by stacking spatio-temporal synchronous modeling modules, which leads to an increase in the model's time complexity. In addition, the spatio-temporal synchronous modeling may ignore global temporal correlation, so TCN and Transformer are often introduced to compensate for the shortcomings of the model in modeling temporal correlation. Therefore, the existing spatio-temporal synchronous models, in a broad sense, are still STNNs.To address these issues, we designed a simple and effective spatio-temporal synchronous traffic flow forecasting model called FasterSTS. FasterSTS does not rely on the important spatio-temporal feature extraction components such as multi-head self-attention mechanism, RNN, and TCN, but only relies on graph convolutional networks to model dynamic spatio-temporal correlations. The specific contributions of FasterSTS are as follows:
\begin{enumerate}
%\item We reform traditional graph convolutional kernel into spatio-temporal synchronous graph convolutional kernel and use a static adaptive embedding to complete the static mapping process of spatio-temporal synchronous convolutional kernel, to capture static temporal correlations during graph convolution.

%\item  To capture dynamic temporal correlations in the mapping process of spatio-temporal synchronous graph convolutional kernel, a dynamic adaptive embedding is obtained through a two-dimensional convolution layer mapping, which completes the dynamic mapping of the convolution kernel to capture dynamic temporal correlations. 
\item
In order to describe the current node, the conventional graph computation approach first collects the information of all other nodes for each node. This operation has a temporal complexity of up to \begin{math}
\mathcal{O}(N2)
\end{math} when working with road network nodes. To drastically cut down on the time complexity of the FasterSTS model, we ingeniously split the conventional graph computing procedure into two main steps:node representation projection and node information aggregation.To effectively collect node information in the node information aggregation step, we employ \begin{math}
N*n
\end{math} (where \begin{math}
n
\end{math} is significantly less than \begin{math}
N
\end{math}) matrices rather than the original \begin{math}
N*N
\end{math} adaptive matrix. The computational effort is significantly reduced by this change. The n-dimensional aggregated data is then projected using $1\times1$ convolutional layer (whose time complexity is \begin{math}
n*N
\end{math}) to produce an N-dimensional node representation in the node representation projection step.

\item 
Based on the use of our proposed novel graph computation approach,we reform traditional graph convolution kernel into spatio-temporal synchronous graph convolution kernel. We achieved the static mapping process of the spatio-temporal synchronous graph convolution kernel using static adaptive embeddings and embedding projection components, which enables static modeling of temporal correlation and static mapping of features during the process of graph convolution. Additionally, we achieved the dynamic mapping process of the spatio-temporal synchronous graph convolution kernel using a two dimensional convolutional layer to obtain dynamic adaptive embeddings, which enables dynamic modeling of temporal correlation and dynamic mapping of features during the process of graph convolution.
%In order to capture dynamically changing spatio features, we introduce spatial context gating into the model, and assign different importance levels to the spatial features extracted by multi-Graph convolution based on the attention mechanism, which improves the model's performance in medium and long-term prediction steps. 
%

\item In order to fully and effectivelty capture spatial correlations, we employ global and local adaptive embeddings, allocating different adaptive graphs to different hidden dimensions. We utilize generated dynamic spatio-temporal synchronous graph convolution kernel to dynamically adjust the obtained spatial features for modeling dynamic spatial correlations.  
%
%The time prediction network based on RNN often pays more attention to the time closer to the current time when predicting. Based on this, we introduced the time transformer module in the model to improve the model's capture of long-distance time correlation ability. Graph convolutions and temporal transformers are combined in parallel in our model, which provides additional contextual information for the model while extracting long-range temporal dependencies.
%
%\item The experimental results show that FasterSTS, with its highly efficient model framework, achieves spatio-temporal synchronous modeling at a much higher efficiency than existing STGNNs methods, while being more time-efficient and outperforming existing STGNNs methods in terms of prediction accuracy.
\end{enumerate}
%

%The remainder of the paper is organized as follows. Section III defines the symbols used in this paper and the traffic flow forecasting problem. Then, the overall structure of our proposed prediction model FasterSTS is presented in Section IV. 
%by modules. 
%Section V presents the experimental results and analysis.
%the comparison experiments and ablation experiments with other models.
%Section VI summarizes and presents future work.
%looks forward to this paper.

% You must have at least 2 lines in the paragraph with the drop letter
% (should never be an issue)
%\hfill August 26, 2015
\section{Related Works}
\subsection{Graph Convolutional Networks}
Traditional convolutional neural networks are usually only suitable for processing regular Euclidean data, while graph convolutional networks extend the convolutional neural networks to general graph structured data. graph convolutional operations have two calculation methods: the spatial graph convolution method \cite{36} and the spectral graph convolution method \cite{37}. The general framework of the spatial graph convolution method mainly has two types, one is the message passing neural network (MPNN) \cite{37}. The other is the mixture model network (MoNet) \cite{39}. Although the spatial graph convolution can be directly calculated in the spatial domain, when processing large-scale graph data, the spatial convolution method requires a large amount of computing resources. The spectral graph convolution designed by the Fourier transform and spectral analysis \cite{40} can use the Chebyshev approximation \cite{41} to reduce the complexity of the Laplacian matrix eigenvalue decomposition, so that GCN can efficiently process spatio-temporal data and achieve excellent performance.
\subsection{GCN-Based Models for Traffic Forecasting}
GCN have expanded the applicability of traditional CNN to graph structured data, solving the problem that CNN cannot handle non-Euclidean data. Methods for traffic flow forecasting that model spatial correlation based on GCN and integrate temporal correlation modeling components have become research hotspots. DCRNN \cite{17} models spatial correlation in traffic flow using a random walk process on directed graphs and combines convolution expansion and GRU to capture temporal dependencies. STGCN \cite{19} combines graph convolution and gated temporal convolution to effectively capture spatial correlations while extracting temporal correlations. Graph WaveNet (GWN) \cite{18} combines graph convolution with extended causal convolution. GWN designs deep one-dimensional convolution expansions with an increasing receptive field that grows exponentially with the number of convolutional layers, thus effectively handling long-term sequence data. Due to the complexity and dynamic nature of traffic flow, ASTGCN \cite{42} uses traditional attention mechanisms to learn an attention matrix. During graph convolution and temporal convolution operations, the learned attention matrix dynamically adjusts the importance of different nodes at the traffic roads and different time steps to capture dynamic spatio-temporal correlations. Transformer can capture dynamic spatio-temporal correlations in temporal data more precisely \cite{15-6}.Compared with traditional attention mechanisms, multi-head self-attention mechanisms can learn multiple sets of different attention weights in parallel to obtain the optimal attention weights for different features, thus further improving the model's ability to model dynamic correlations. STTNs \cite{21-1} simultaneously includes spatial Transformer and temporal transformer, with the spatial transformer used to capture dynamic spatial correlations and the temporal transformer used to capture long-range dynamic temporal dependencies. Traffic-Transformer \cite{43} built a special encoding and feature embedding to solve the problem of incompatibility between the transformer and traffic flow data. In addition, the original encoder and decoder structure was improved to a global encoder and a global-local decoder to model global and local spatio-temporal correlations. ASTGNN \cite{44} designed a temporal trend-aware multi-head self-attention to solve the problem of insufficient temporal perception in traditional multi-head self-attention mechanisms.
\subsection{Spatio-Temporal synchronous Models for Traffic Forecasting}
Compared with the spatio-temporal asynchronous modeling methods mentioned above, STSGCN \cite{34} and STFGNN \cite{35} construct a spatio-temporal graph by  temporal graph and topological graph, and use the spatio-temporal graph to extract heterogeneous spatio-temporal correlations between adjacent time steps. STGODE \cite{47} builds a deep network based on tensor's ordinary differential equation to extract spatio-temporal correlations synchronously. However, existing spatio-temporal synchronous methods often obtain a high-time-complexity spatio-temporal graph by concatenating the temporal graph and spatial graph, making it difficult to model spatio-temporal correlations dynamically. Additionally, TCN and Transformer are often introduced in spatio-temporal synchronous modeling to compensate for the shortcomings of the model in modeling global and dynamic temporal correlations, which further increases the complexity of the model.
Compared with the spatio-temporal asynchronous modeling methods mentioned above, STSGCN \cite{34} and STFGNN \cite{35} construct a spatio-temporal graph by  temporal graph and topological graph, and use the spatio-temporal graph to extract heterogeneous spatio-temporal correlations between adjacent time steps. STGODE \cite{47} builds a deep network based on tensor's ordinary differential equation to extract spatio-temporal correlations synchronously. However, existing spatio-temporal synchronous methods often obtain a high-time-complexity spatio-temporal graph by concatenating the temporal graph and spatial graph, making it difficult to model spatio-temporal correlations dynamically. Additionally, TCN and Transformer are often introduced in spatio-temporal synchronous modeling to compensate for the shortcomings of the model in modeling global and dynamic temporal correlations, which further increases the complexity of the model.
\section{Spatio-Temporal
Synchronous Graph Convolutional Networks}
%
% \subsection{Problem description}
%
\subsection{Preliminaries}
As a typical time series forecasting task, the traffic flow forecasting is to use the traffic flow at historical moments collected by sensors to predict the traffic flow at future moments. In order to facilitate the description of the traffic flow forecasting problem, the road network was defined as a weighted graph $\mathcal{G}=(\mathcal{V}, \mathcal{E},\mathcal{A})$.
%(referred to as "road network"), 
%
Herein, each vertex $v_i \in \mathcal{V}$ represents a traffic node in the road network. Each edge $e_{ij} \in \mathcal{E}$ denotes the road segment connecting traffic nodes $v_i$ and $v_j$. The weight $a_{ij} \in \mathcal{A}$ reflects the spatial correlation between traffic nodes $v_i$ and $v_j$. 
%In this article, how to compose the image is one of the keys to our method. We will do this from two perspectives: road network topology (represented by $\mathcal{G}_a)$ and distance correlation (represented by $\mathcal{G}_d)$. Construct spatial correlation relationships between road networks. 
%
%The traffic flow prediction task is a time series prediction task. Traffic flow prediction can be expressed as learning an optimal mapping function $\mathcal{F}_p()$, which combines the traffic flow data collected by the sensor for the past T historical moments $[X^{t-T},. ..,X^{t-1}]$, mapped to $\tau$ future moments traffic flow data values $[X^{t+1},...,X^{t+\tau}]$:}
%
In fact, as defined in Eq.\ref{eq001}, the essence of traffic flow forecasting task is to learn an optimal traffic flow model $\mathcal{F}_\mathit{p}()$ that maps the $T$ historical moments
traffic flow data $[X^{\mathit{t-T+1}},. ..,X^{\mathit{t}}]$ to the $\tau$ future moments traffic flow data $[\hat P^{\mathit{t+1}},...,\hat P^{\mathit{t+\tau}}]$. 
%
%*******************************%
%
\begin{equation}\label{eq001}
[{\hat P}^{\mathit{t+1}},...,{\hat P}^{\mathit{t+\tau}}] = \mathcal{F}_\mathit{p}\ ([{X}^{\mathit{t-T+1}} ,...,{X}^{\mathit{t}}],\mathcal{G})
\end{equation}
%
%*******************************%
%
where $\mathit{p}$ denotes the set of hyperparameters of the traffic flow forecasting model $\mathcal{F}_p()$. ${X}^\mathit{t} = [{X}_\mathit{1}^\mathit{t},{X}_\mathit{2}^{\mathit{t}},...,{X}_\mathit{n}^\mathit{t}] \in R^\mathit{N\times C}$ represents the set of traffic flows of the $N$ traffic nodes in the road network at time $t$.
% needed in second column of first page if using \IEEEpubid
%\IEEEpubidadjcol
%\subsubsection{Subsubsection Heading Here}
%Subsubsection text here.
%
%*****************************%
%

%The graph convolution operation in this paper is carried out on the above three graphs respectively. 
%
%the graph convolution operation on topological map, distance map and traffic pattern map can be recorded as $\mathbit{\Theta} \begin\ \ast _{\mathcal{G}_r}x$ , $\mathbit{\Theta} \begin\ \ast _{\mathcal{G}_d}x$ and $\mathbit{\Theta} \begin\ \ast _{\mathcal{G}_p}x$.
%
%*****************************%
%
\begin{figure*}
     \centering
     \begin{subfigure}[b]{0.48\textwidth}
         \centering
             \includegraphics[width=0.75\textwidth]{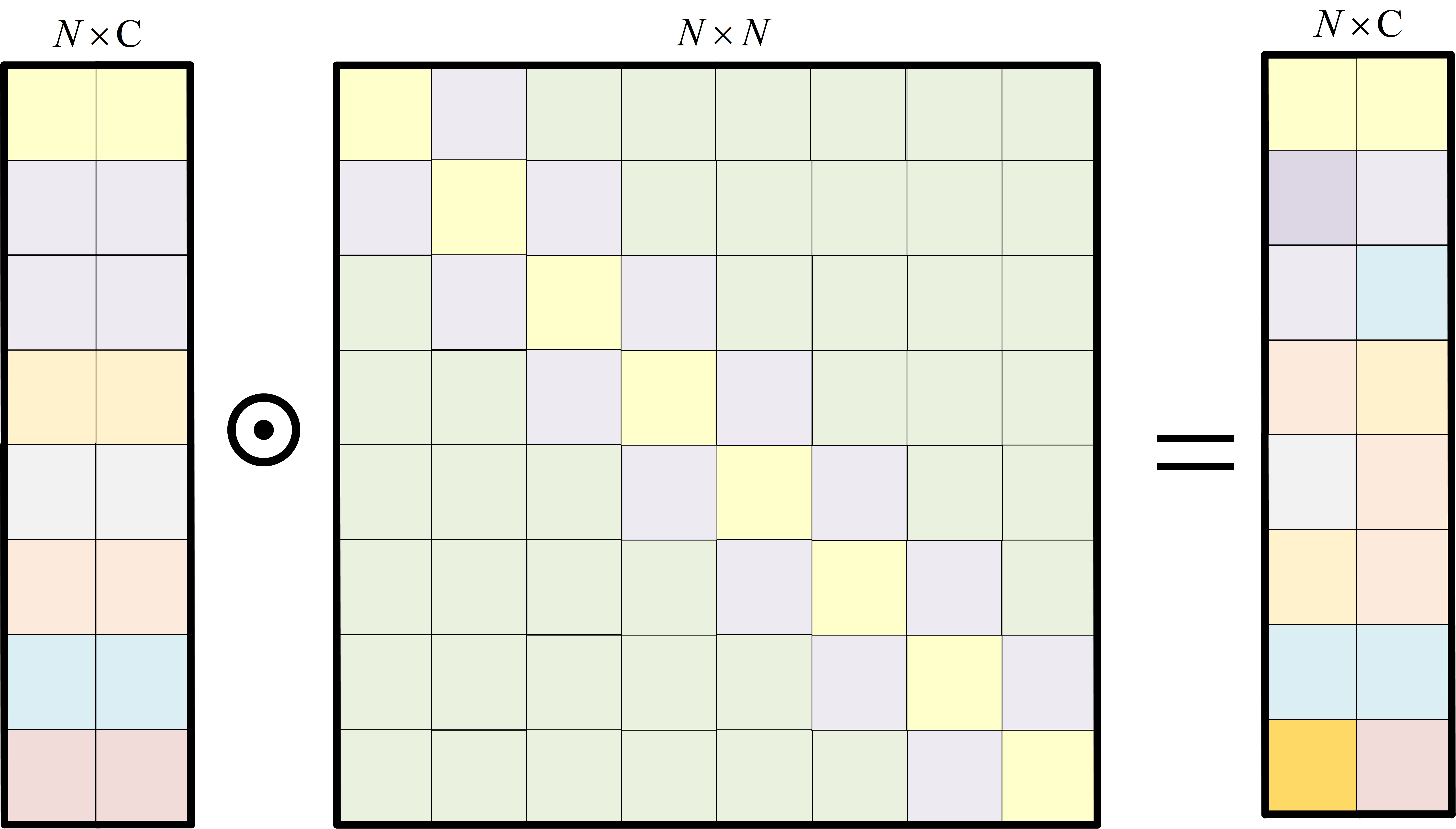}
         \caption{The process of traditional graph computation}
         \label{fig:1a}
     \end{subfigure}
     \hfill
     \begin{subfigure}[b]{0.5\textwidth}
         \centering
         \includegraphics[width=\textwidth]{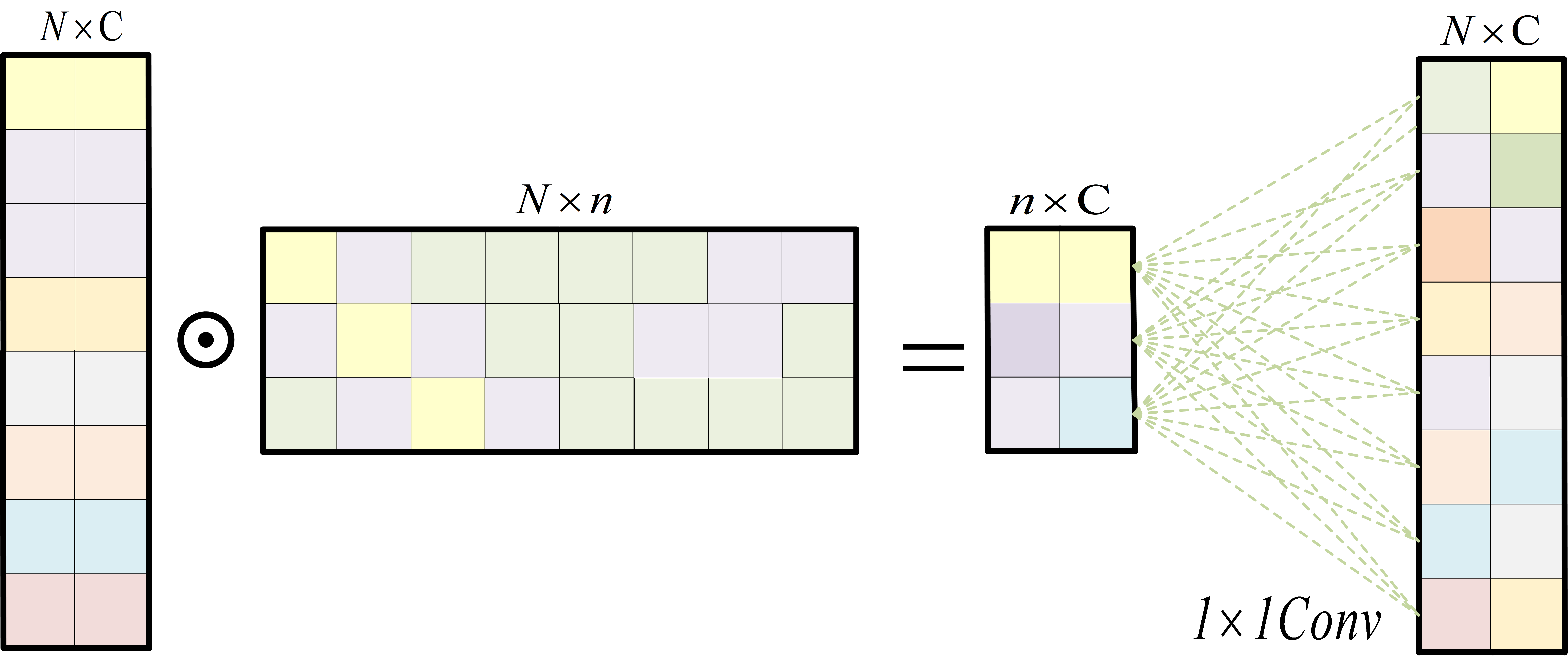}
         \caption{The process of Faster graph computation}
         \label{fig:1b}
     \end{subfigure}
     \hfill
     \caption{The process of graph computation with different methods}
        \label{fig:1}
\end{figure*}
\subsection{Overall Architecture}
As shown in Fig.\ref{fig:arch}, we have proposed an  efficient spatio-temporal synchronous traffic flow forecasting model that extracts complex dynamic spatio-temporal correlations. It includes an input layer, a number of stacked STSGCL, and a fusion layer. The input layer maps spatio-temporal features into a high-dimensional space to enhance the model's expressive ability through  a fully connected layer. The key of the STSGCL is the spatio-temporal synchronous graph convolution kernel, which is obtained by the Hadamard product of the static adaptive embedding and the dynamic adaptive embedding to extract static and dynamic spatio-temporal correlations and complete the feature mapping. In addition, We have developed an adaptive graph specifically for the spatio-temporal synchronous convolution kernel, which can effectively model both static and dynamic spatial correlations. Finally, to accelerate the convergence speed and stability of the model, residual connections and normalization are employed in each STSGCL. The fusion layer aggregates different granularity spatio-temporal features and uses two fully connected layers to process the predicted output.

%This is sufficient to fuse the output of the multi-graph convolution module and make a better multi-step prediction. 
The following will introduce the working principle of each module of the model.
%and the role of this module in traffic flow prediction.
%
%*******************************%
%
%
%*******************************%
%
%
%*******************************%
%
\subsection{Input Embedding Layer}
The inputs are projected into a high-dimensional space using a fully-connected layer in the top layer of our proposed FasterSTS, which also incorporates spatio-temporal location coding and information from the previous day’s temporal cycles. This improves the FasterSTS model's ability to represent the traffic flow's periodicity and spatio-temporal features. We use adaptive embedding \begin{math}
X_{pe}
\end{math} for spatio-temporal location coding. In terms of periodicity, we regard each week as having seven distinct time slots and divide the time of day into 1440 time slots.These time slot data are then embedded using embedding functions to obtain corresponding representations of periodic features \begin{math}
X_{w}
\end{math} and \begin{math}
X_{d}
\end{math}, denoted as The input \begin{math}
X
\end{math} of FasterSTS was then obtained by fusing these two periodic features with spatiotemporal location coding.Eq.(\ref{eq:002}) provides a definition for the input embedding procedure.
\begin{equation}\label{eq:002}
{X} = X_{in}  + X_{w} + X_{d} + X_{pe}
\end{equation}
Here in, $X_{in}$ is the original traffic flow data to FasterSTS. $X$ is the high-dimensional representation of $X_{in}$ after spatio-temporal location coding.
\subsection{Faster graph computation}
In the process of graph computation, to describe the current node, traditional graph computation methods would first collect information from all other nodes for each node. When dealing with road network nodes, this operation incurs a time complexity as high as \begin{math}
\mathcal{O}(N^2)
\end{math}. To drastically reduce the time complexity of the FasterSTS model, we ingeniously decompose the traditional graph computation process into two main stages: node representation projection and node information aggregation.The comparison of the computation processes between traditional graph computation and fast graph computation is shown in the Fig.\ref{fig:1}. This approach, which we refer to as fast graph computation, can also be extended to some generalized graph computations.For example, the fully connected mapping process.
\begin{figure*}[ht!]
\centering
  \includegraphics[width=.99\linewidth]{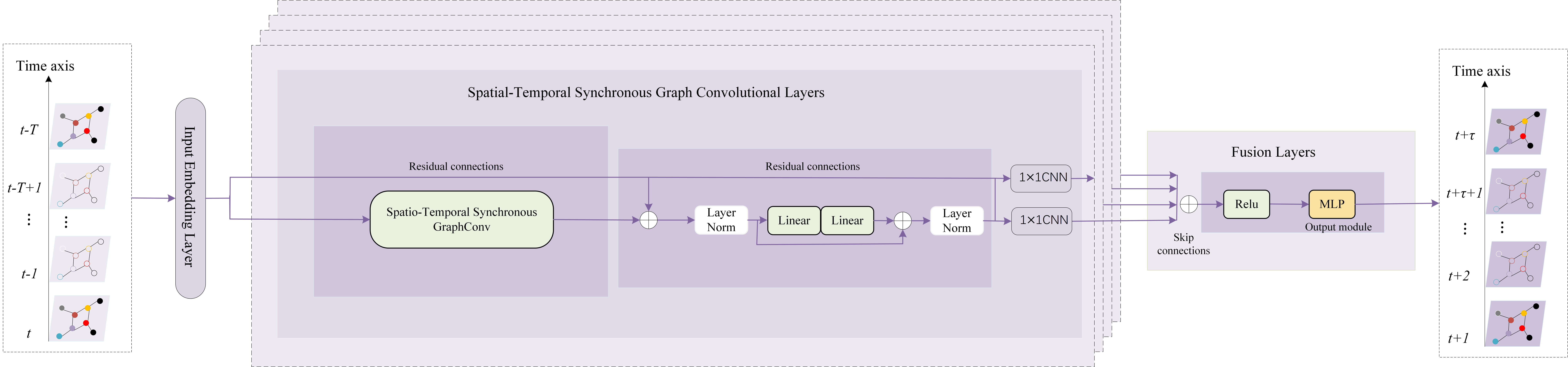}
  \caption{The architecture of FasterSTS }
  \label{fig:arch}
\end{figure*}
\subsubsection{Node Information Aggregation}
In the process of node information aggregation, we employ an adaptive matrix \begin{math}
E \in R^{N \times n}
\end{math} (where \begin{math}
n
\end{math} is significantly less than \begin{math}
N
\end{math}) with a time complexity of \begin{math}
nN
\end{math},instead of the original \begin{math}
N \times N
\end{math} adaptive matrix, to efficiently collect node information. This change significantly reduces the computational load. If the input is \begin{math}
X_F \in R^{T \times N \times C}
\end{math}, the projection process can be defined as Eq.(\ref{eq003}):
\begin{equation}\label{eq003}
\begin{split}  
& A(i,j)\  = \left[Softmax(E_{:,j}) \right]_i\\
&\hat X_F= A  \odot X_F
\end{split}
\end{equation}
Here in,$\odot$ represents the Hadamard product.$\hat X_F \in R^{T \times n \times C}$ is the output after performing fast graph computation on $A$ and $X_F$.
\subsubsection{Node Representation Projection}
In the node representation projection process, we utilize 1$\times$1 convolution layer (with a time complexity of \begin{math}
nN
\end{math}) to project the $n$-dimensional aggregated data, thereby generating an $N$-dimensional node representation. The specific process can be defined as Eq.(\ref{eq004}):
\begin{equation}\label{eq004}
\tilde X_F = {C_F}\hat X_F + {c_F}
\end{equation}
Among them,$\tilde X_F \in R^{T \times N \times C}$ is the output after performing node representation projection, while ${C_F}$ represents convolution operations, ${c_F}$ is learnable parameters.
\subsection{Spatio-Temporal synchronous Graph Convolutional Layers}
%The graph generation process is one of the keys to successfully building multi-graph convolutional models. 
%Whether the generated graph can accurately reflect the spatio correlation between the traffic nodes in the network directly affects the effect of the graph convolution operation.
%
As shown in the Fig.\ref{fig:arch}, FasterSTS is composed of multiple spatio-temporal synchronous graph convolutional layers. The spatio-temporal synchronous graph convolution is the key component of FasterSTS, denoted as $STG()$. To model the temporal correlation in the graph convolution process, $STG()$ reforms the traditional graph convolution kernel $\Theta  \in R^{{d_{in}} \times {d_{out}}}$ into a spatio-temporal graph convolution kernel $\Psi  \in R{}^{T{d_{in}} \times T{d_{out}}}$.

\subsubsection{Spatio-Temporal Synchronous Graph Convolutional Networks}
\begin{figure}[ht!]
\centering
  \includegraphics[width=.99\linewidth]{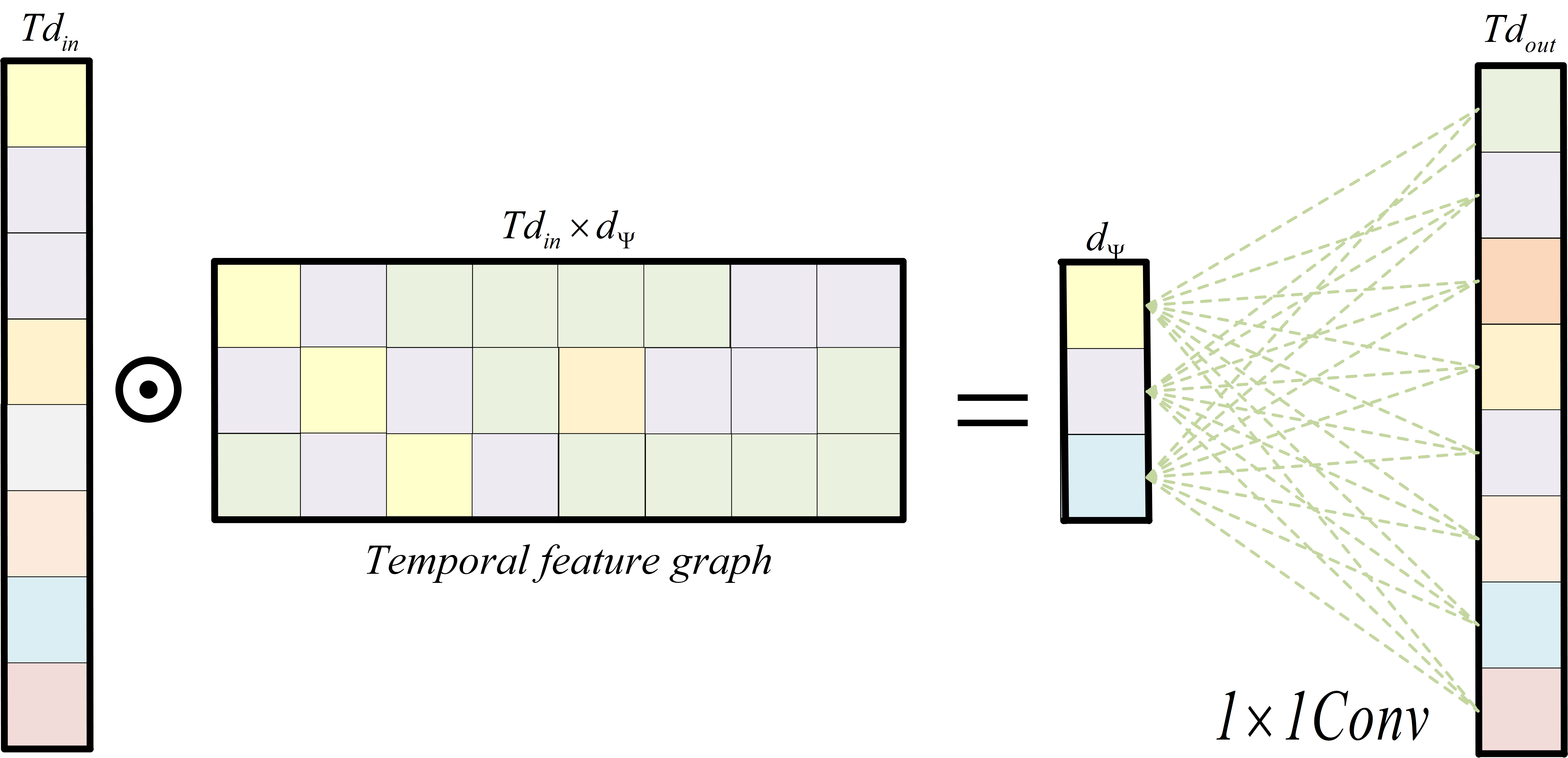}
  \caption{The computation process of the spatio-temporal synchronous graph convolution kernel}
  \label{fig:6}
\end{figure}

For the spatially related graph $\mathcal{G}$, GCN uses the corresponding adjacency matrix $A$ to model the spatio correlation between different traffic nodes. In the temporal forecasting task, existing research uses static graph convolutional kernels to map features. Based on the 1stCheb-net approximation algorithm \cite{23}, the specific calculation process of the graph convolution between the traffic flow $X$ and the graph $\mathcal{G}$ can be defined as Eq.(\ref{eq005}):
\begin{equation}\label{eq005}
\begin{split}
&SG(X) = \Theta { * _g}X = \Theta ({I_N} + {D^{ - 1}}A)X \\
&=\Theta ({\tilde D^{ - \frac{1}{2}}}\tilde A{\tilde D^{ - \frac{1}{2}}}X)
\end{split}
\end{equation}
Where $X\in R^{T \times N \times {d_{in}}}$ represents the input signal, $\Theta  \in R{}^{{d_{in}} \times {d_{out}}}$ is the convolution kernel in the graph convolution, and $I$ is the identity matrix. $D$ is the degree matrix of the nodes, which satisfies ${D_{ij}} = \sum\limits_j {{A_{ij}}}$.
The graph convolution kernel in traditional graph convolution cannot capture temporal correlation during the graph convolution process, ignoring the spatio-temporal heterogeneity in traffic flow data. To realize the spatio-temporal synchronous modeling of graph convolution, we propose the spatio-temporal synchronous graph convolution based on fast graph computation,which models temporal correlations and feature mappings in the form of a temporal feature graph.Its computation process is shown in Fig.\ref{fig:6}.The spatio-temporal synchronous graph convolution can be expressed as Eq.(\ref{eq006}):
\begin{equation}\label{eq006}
\begin{split}
&{{\hat X}} = {\tilde D^{ - \frac{1}{2}}}\tilde A{\tilde D^{ - \frac{1}{2}}}{{X}}\\
&{{\tilde X}} = {{R}}(\hat {X})\\
&\overline{X}  = \Psi { * _g}{{\tilde X}} = \Psi \odot {R}(({I_N} + {D^{ - 1}}A){{X}}) = \Psi  \odot {R}({{\hat X}})\\
&STG({{\overline X}}) = {C_\Psi} \cdot \overline{X} + {c_\Psi}
\end{split}
\end{equation}
Where $X\in R^{N \times T \times {d_{in}}}$ represents the input signal,$\overline X\in R^{N \times d_\Psi}$ represents the input signal, $\hat X \in R^{N \times T \times {d_{in}}}$ represents the output of the matrix multiplication between ${X}$ and ${\tilde D^{ - \frac{1}{2}}}\tilde A{\tilde D^{ - \frac{1}{2}}}$, ${\tilde X} \in {R^{N \times T{d_{in}}}}$ represents the output of $\hat X \in R^{N \times T \times {d_{in}}}$ after the dimension transformation,${C_\Psi}$ represents convolution operation, ${c_\Psi}$ is learnable parameter. ${{R}}\left( \right)$ represents the operation of dimension transformation  that transforms the feature dimension $N \times T \times {d_{in}}$ \ to the dimension $N \times T{d_{in}}$, $\Psi  \in R^{T{d_{in}} \times d_\Psi}$ is the convolution kernel of the spatio-temporal synchronous graph convolution operation, and $I$ is the identity matrix. $D$ is the degree matrix that satisfies the equation ${D_{ij}} = \sum\limits_j {{A_{ij}}}$. The output of the spatio-temporal synchronous graph convolution operation is $STG({{\overline X}}) \in {R^{N \times T{d_{out}}}}$.
%If the generated graph can not accurately reflect the spatio correlation between the traffic nodes in the network, the performance of the graph convolution operation will be restricted.
%
%If the graphs we generate cannot effectively encode the spatio correlation between traffic nodes, and at the same time, the larger the number of graphs, the better the performance of the model will not necessarily be. If the generated graphs are inappropriately numeric, it will not contribute to the learning of model parameters and may even have a negative impact on the model's predictive performance. On this basis, two methods for constructing the relationship diagram between traffic nodes are proposed: topology graph and distance graph.In order to effectively modeling the spatio correlations between different traffic nodes of the road  network, a topology graph and a distance graph 
%
%Although different graphs can be constructed from different dimensions to model the spatio correlation between different traffic nodes in the road network, the more graphs constructed is not the better. Therefore, as done in many of the existing works, this paper defines a topology graph and a distance graph to describe the spatio correlation between different nodes in the network.
%
%

In order to fully capture the temporal correlation in the graph convolution process, we divide the graph convolution kernel in the spatio-temporal synchronous graph convolution into dynamic and static parts to realize static and dynamic modeling of temporal correlation.
\begin{itemize}
  \item The static part of the spatio-temporal synchronous graph convolution kernel
\end{itemize}
To enable temporal correlation to be captured during the convolution process, we reform the traditional graph convolution kernel $\Theta  \in R{}^{{d_{in}} \times {d_{out}}}$ into a spatio-temporal graph convolution kernel $\Psi  \in R{}^{T{d_{in}} \times T{d_{out}}}$. Our inspiration comes from HTVGNN\cite{15-2}, and we construct a spatio-temporal synchronous graph convolution kernel $\Psi $ to capture static and dynamic temporal correlation between different time steps. First,We use a learnable embedding $\Upsilon  \in {R^{T \times L \times d}}$ to adaptively learn static temporal correlation and complete the feature mapping.
%
%Therefore, links of roads that are directly connected in a road network will be assigned a higher weight association. 
%Fig.\ref{fig:sub4} and Table.\ref{fig:sub5} respectively give an example of the weight matrix of a road network and its topology. 
\begin{itemize}
  \item Reinforcement Learning Component
\end{itemize}
In order to better learn the temporal correlation between different time steps in traffic flow and complete the feature mapping, we designed the Reinforcement Learning(RL) Component. The RL component provides a low-dimensional representation for $\Upsilon  \in {R^{T \times L \times d}}$ embedding and captures hidden temporal dependencies. In the RL component, we use two fully connected layers to project $\hat \Upsilon  \in {R^{TL \times d}}$ to an embedding $\tilde \Upsilon  \in {R^{T{d_{in}} \times d}}$ of the same dimension as the  spatio-temporal synchronous convolutional kernel $\Psi $, which is used to model static temporal correlation.Its form is as Eq.(\ref{eq007}):
\begin{equation}\label{eq007}
\begin{split}
&\hat \Upsilon  = {{R}}(\Upsilon )\\
&\tilde \Upsilon  = \Omega (\hat \Upsilon\cdot{W_2} + b_2)\cdot{W_3+b_3}\\
%&\Omega (x) = \left\{ \begin{array}{l}
%x,x > {\delta _\Omega }\\
%0,otherwise
%\end{array} \right.
\end{split}
\end{equation}
Where ${W_2}$ , ${W_3}$ and $b_2$ , $b_3$ are learnable weight parameters, $\hat \Upsilon  \in {R^{TL \times d}}$ represents the output of the dimension transformation performed by $\Upsilon  \in R^{T \times L \times d}$. ${R}\left( \right)$ represents the transformation operation that converts the feature dimension $T \times L \times d$ to the dimension $TL \times d$. $\Omega ()$ represents the ReLU activation function, which is applied to enhance the anti-interference ability.
%
%
%***************************************%

%
\begin{itemize}
  \item The dynamic part of the spatio-temporal synchronous graph convolution kernel
\end{itemize}
To enable the spatio-temporal convolutional kernel $\Psi$ to capture dynamic temporal correlations between different time steps while reducing model complexity, we first convert $\hat X \in R^{N \times T \times {d_{in}}}$ to $\tilde X' \in R^{N \times T{d_{in}}}$ based on fast graph operations. Then, we use a 1×1 convolutional layer to transform $\tilde X'$ into an intermediate vector $X' \in R^{T{d_{in}  \times d}}$. To obtain the static part of $\Psi$, we perform element-wise multiplication between $X'$ and $\Upsilon$. This process is shown in Eq.(\ref{eq008}).
%In order to enable the spatio-temporal convolutional kernel $\Psi$ to capture the dynamic temporal correlation between different time steps while reducing the model complexity. First, we use a 1$\times$1 convolution layer to reduce $\hat X \in R^{N \times T \times {d_{in}}}$ to a midddle vector $X' \in R^{N \times T \times {d_{out}}}$ and then transform it to $\tilde X' \in R^{N \times T{d_{out}}}$. To obtain the dynamic part of $\Psi$, we use another 1$\times$1 convolution layer to reduce $\tilde X' \in R{}^{N \times T{d_{out}}}$ to a vector $\tilde {\rm Z} \in R{}^{d \times T{d_{out}}}$ based on $\tilde X'$. The process is as Eq.(\ref{eq005}):
\begin{equation}\label{eq008}
\begin{split}
&{\rm{\tilde X'}} = {{R}}(\hat X)\\
&Z =  X' \cdot \tilde \Upsilon\\
\end{split}
\end{equation}
Among them, ${{R}}()$ represents the transformation operation that converts feature dimension $N \times T \times {d_{in}}$ to the dimension $N \times T{d_{in}}$, while ${C_1}$ and ${C_2}$ represent convolution operations, ${c_1}$ and ${c_2}$ are learnable parameters.

We obtain the spatio-temporal synchronous graph convolution kernel by normalizing it using the Softmax() function. The process is as Eq.(\ref{eq009}):
\begin{equation}\label{eq009}
\Psi  = \left[Softmax(Z_{:,j}) \right]_i
\end{equation}

\begin{itemize}
  \item Residual connections and normalization
\end{itemize}
To prevent gradient vanishing and explosion, and accelerate the convergence rate of the STSGCL\cite{15-6} \cite{46}, residual connections and layer normalization operations are employed into each STSGCL of the FasterSTS. The specific form is as Eq.(\ref{eq0091}):
\begin{equation}\label{eq0091}
\begin{split}
&{{\tilde X}} = Layernorm(STG({{\overline X}}) + {{X}})\\
&{{\bar X}} = Layernorm(Relu({{\tilde X}}\cdot{W_4 + b_4})\cdot{W_5} + b_5 +  {{\tilde X}}))
\end{split}
\end{equation}
where ${W_4}$ , ${W_5}$ and $b_4$ , $b_5$ are learnable parameters,$Relu()$ is the activation function,and $Layernorm()$ is layer normalization.

Residual connections help promote the transmission of input information to lower layers, thus alleviating the problem of difficulty in transmitting information in deep networks. As shown in the figure, when transmitting information between two layers of the model, we further adopted residual connections, which are defined as Eq.(\ref{eq011}):
\begin{equation}\label{eq011}
{{\bar X}^{(l)}} = {\bar X}\cdot{W_6} + b_6  +  {\bar X}
\end{equation}
Where $W_6$ and $b_6$ are learnable weight parameters, $\bar X$ represents the output after residual connection and normalization, and ${\bar X}^{(l)}$ refers to the output of each STSGCL after further residual connection.
\begin{itemize}
  \item Adaptive graph generation
\end{itemize}
Graph convolution is a process of aggregating information on a graph, and the quality of the graph directly affects the performance of graph convolution. Since the predefined graph is generated by domain experts based on prior knowledge, the predefined graph often cannot fully reflect the real road network. Therefore, we use adaptive graphs based on fast graph computation, in the STSGCL to capture the spatio correlation of the road network. Considering that the spatio-temporal synchronous graph convolution kernel in our proposed spatio-temporal synchronous graph convolution is dynamic, we use two learnable embedding to learn different static adaptive graphs for each hidden dimension $D$ \cite{15-2}, and then use dynamic $\Psi$ to dynamically adjust ${{\hat X}} = {\tilde D^{ - \frac{1}{2}}}\tilde A{\tilde D^{ - \frac{1}{2}}}{{X}}$ by means of likewise gated operation to capture the dynamic spatial correlation of the road network. To enable $\Psi$ to dynamically adjust the static adaptive graph while modeling dynamic temporal correlation, different hidden dimensions are assigned different adaptive graphs in the STSGCL, and the static adaptive adjacency matrix $A_l$ of the $l$-$th$ hidden dimension is defined as Eq.(\ref{eq012}):
\begin{equation}\label{eq012}
\begin{split}
&{E_l} = E + {e_l}\\
&{A_l} = \left[Softmax({(E_l)}_{:,j}) \right]_i 
\end{split}
\end{equation}
Specifically, $E \in R^{N \times d_e}$ is a trainable global node embedding, ${e_l} \in R^{l \times d_e}$ is a trainable local node embedding at the $l$-$th$ hidden dimension,$Softmax()$ is the activation function.
Based on fast graph operations and adaptive graphs, ${{\hat X}} = {\tilde D^{ - \frac{1}{2}}}\tilde A{\tilde D^{ - \frac{1}{2}}}{{X}}$ can be defined as Eq.(\ref{eq013}):

\begin{equation}\label{eq013}
\hat X_{agg} = A \odot X
\end{equation}
Here in,$\hat X \in R^{d_e \times T \times C}$ is the output after node information aggregation in fast graph computation.

Based on fast graph operations, a 1×1 convolutional layer is utilized to perform node representation projection on $\hat X$.The process is as  Eq.(\ref{eq014}):

\begin{equation}\label{eq014}
\hat X = {C_\mathcal{G}} \cdot \hat X_{agg} + {c_\mathcal{G}}
\end{equation}
Here in,$\hat X \in R^{N \times T \times C}$ is the output after node representation projection in fast graph computation. ${C_\mathcal{G}}$ represent convolution operations, ${c_\mathcal{G}}$ is learnable parameter.
\subsection{Fusion layer}
\subsubsection{Skip connection}
After capturing the spatio-temporal features in STSGCL, we fuse the spatio-temporal features ${{{\bar X}}^{(l)}} \in {R^{N \times T \times C}}$ extracted from each STSGCL. To fully utilize the spatio-temporal features extracted by different STSGCL, we use 1$\times$1 convolution layers without sharing weights to aggregate the spatio-temporal features extracted from each STSGCL, thus enhancing the model's expressive ability. Its definition is as Eq.(\ref{eq015}):
\begin{equation}\label{eq015}
    {{\bar X}_{com}} = \sum\limits_{l = 0}^L {{\Theta _l} * {{{\bar X}}^{(l)}}}
\end{equation}
Where ${\bar X}_{com}$ represents the spatio-temporal features combined through skip connections, while $\Theta _l$ denotes independent convolutional kernel.
%After capturing spatio-temporal features from each STSGCL, we fused the spatio-temporal features extracted from each STSGCL. To fully preserve the spatio-temporal features extracted from different STSGCL, we aggregated the spatio-temporal features extracted from each STSGCL, and further adjusted the dependency between sequences using linear transformation and residual connection. The specific format is as Eq.(\ref{eq009}):
%\begin{equation}\label{eq009}
%\begin{split} 
%&{{{\bar X}}_{agg}} = \sum\limits_{l = 0}^L {{{{{\bar X}}}^{(l)}}}\\
%&{{{X}}_{out}} = Layernorm({W_5}({W_6}{{{\bar X}}_{agg}}) + {{{\bar %X}}_{agg}})\\
%\end{split}
%\end{equation}
%where ${W_5}$ and ${W_6}$ are learnable parameters, and $L$ represents the number of layers in the STSGL.
%
\subsubsection{Output module}
In the output module, we use two fully connected layers with non-shared weights to perform multi-step forecasting, the process of which can be expressed as Eq.(\ref{eq016}):
\begin{equation}\label{eq016}
{\hat P} = \Gamma({\bar X}{_{com}\cdot{W_7}} + {b_7})\cdot{W_8} + {b_8}
\end{equation}
Among them, ${\bar X}_{com}$ denotes the output of the skip connections. $\Gamma\left(\right)$ is the ReLU activation function.${W_7}$ , ${W_8}$ and ${b_7}$ , ${b_8}$ are trainable parameters, and $\hat P \in {R^{T \times N \times C}}$ is the multi-step forecasting output.
\subsection{Loss function}
Because there are missing and abnormal sampling values in the collected traffic flow data, the MAE loss function has robustness and insensitivity to outliers. Therefore, we choose the MAE as the loss function. Its expression is as Eq.(\ref{eq017}):
\begin{equation}\label{eq017}
L({P},{\hat P}) = |{P} - {\hat P}|
\end{equation}
\begin{table*}[htbp]
\renewcommand\arraystretch{1.2}%row height\textbf{}
\centering
\caption{The details of  the datasets}
\begin{tabular}{ccccccccccccccccccccc}     \hline
\multicolumn{4}{c}{\multirow{1}*{Dataset}}     & \multicolumn{4}{c}{\multirow{1}*{PEMS03}}  & \multicolumn{4}{c}{\multirow{1}*{PEMS04}}  & \multicolumn{4}{c}{\multirow{1}*{PEMS07}}& \multicolumn{4}{c}{\multirow{1}*{PEMS08}}\\ \hline
  \multicolumn{4}{c}{\# Nodes}                             &  \multicolumn{4}{c}{358}    &    \multicolumn{4}{c}{307} &  \multicolumn{4}{c}{883}    &    \multicolumn{4}{c}{170}  \\
  \multicolumn{4}{c}{\# Edges}                             &  \multicolumn{4}{c}{866}    &    \multicolumn{4}{c}{340} &  \multicolumn{4}{c}{340}    &    \multicolumn{4}{c}{277}  \\
            \multicolumn{4}{c}{Time steps}                             &  \multicolumn{4}{c}{ 26208}   &  \multicolumn{4}{c}{16992}  &  \multicolumn{4}{c}{28224}   &  \multicolumn{4}{c}{17856}     \\
      \multicolumn{4}{c}{Time span}                             &  \multicolumn{4}{c}{ 2018/9/1–2018/11/30}   &  \multicolumn{4}{c}{ 2018/1/1–2018/2/28}  &  \multicolumn{4}{c}{ 2017/5/1–2017/8/31}   &  \multicolumn{4}{c}{ 2016/7/1–2016/8/31}     \\

        \multicolumn{4}{c}{Missing ratio}                             &  \multicolumn{4}{c}{ 0.672\%}   &  \multicolumn{4}{c}{3.182\%}  &  \multicolumn{4}{c}{0.452\%}   &  \multicolumn{4}{c}{0.696\%}  \\
   \multicolumn{4}{c}{ Time interval}                             &  \multicolumn{16}{c}{5 min}\\
   \multicolumn{4}{c}{daily range}                             &  \multicolumn{16}{c}{0:00-24:00}\\
       \hline

\end{tabular}\label{tb:1}
\end{table*}
Among them, ${P}$ represents the observed traffic flow, ${\hat P}$ represents the predicted traffic flow.
\subsection{Algorithmic details}
To provide a more accurate description of the algorithm flow of the FasterSTS model we propose, we present the training process of the FasterSTS model using Algorithm \ref{algo:1}.
% Reminder: the "draftcls" or "draftclsnofoot", not "draft", class
% option should be used if it is desired that the figures are to be
% displayed while in draft mode.
%
\floatname{algorithm}{Algorithm 1:}
\begin{algorithm}[t]
    \caption{algorithm:1}
	\label{algo:1}
  \renewcommand{\algorithmicrequire}{\textbf{Input:}}
\renewcommand{\algorithmicensure}{\textbf{Output:}}
\renewcommand{\thealgorithm}{}
 \caption{Training Process of The FasterSTS Model.}
    \begin{algorithmic}[1]
    	\REQUIRE  historical traffic flow data of the previous $T$ moments:${X}  =  [{X}^{\mathit{t-T+1}},. ..,{X}^{\mathit{t}}]$;Adaptive adjacency matrix:  ${A}$.
     %, a learnable embedding $\Upsilon  \in {R^{T \times L \times d}}$.

                           \ENSURE Multi-step  traffic flow  forecasting data $[\hat P_{t+1},\hat P_{t+2},...,\hat P_{t+\tau}] $.
    \STATE Initialize the FasterSTS parameters randomly\\
    \STATE construct training  set from the historical traffic flow data $D_{train} \gets \left (X_{train},P_{train} \right)$\\
  \FOR{0 \textless \textit{epoch} \textless  \textit{epochs}}
                \FOR{${X}$ \textit{in} $D_{train}$ }
                    \STATE ${X} \gets [{X}]$ using the input layer
                        \FOR{0 $\textless$ \textit{l} $\textless$ \textit{L}}    
            \STATE  $\tilde \Upsilon \gets \Upsilon$ through Eq.$\ref{eq004}$
             \STATE ${{\hat X}} \gets {X}$ through Eq.$\ref{eq013}$ and Eq.$\ref{eq014}$
            \STATE  $\Psi \gets [\hat X,\tilde \Upsilon]$ through Eq. $\ref{eq008}$ and Eq. $\ref{eq009}$
            \STATE $STG({{X}}) \gets [{X},A,\Psi]$ through Eq. $\ref{eq006}$
            \STATE ${{\bar X}} \gets [STG({{X}}),{X}]$ through Eq. \ref{eq0091}
                    \ENDFOR
           \STATE ${X}_{com} \gets [\bar X^{\mathit{0}},...,\bar X^{\mathit{L}}]$ through Eq. \ref{eq015}
           \STATE ${\hat P} \gets [{X_{com}}]$ through Eq.$\ref{eq016}$
           \STATE the parameters are updated using gradient descent to minimize the Eq.$\ref{eq011}$.
                  \ENDFOR
                  \ENDFOR
    \end{algorithmic}
\end{algorithm}
%
%***************************************************************%
%
\section{Experimental Results and Analysis}
%
%***************************************************************%
%
%\subsection{Model Traffic Prediction Algorithm Process}
%To enhance the clarity of our model's algorithm flow description, we have devised an algorithmic flowchart (Fig.\ref{4}) to illustrate the sequential steps of our model.
%
%*********************************
%
%\begin{figure*}[ht！]
%\centering
%\includegraphics[width=0.3\linewidth,height=7cm]%{process.png}
%\caption{Seq2Seq Prediction }
%\label{fig1}\label{4}
%\end{figure*}
%
%*********************************
%
\subsection{Datasets}
Building on the work of previous researchers \cite{15-2,1801}, we evaluated the single-step and multi-step forecasting performance of FasterSTS on four real-world public traffic flow datasets PEMS03, PEMS04, PEMS07 and PEMS08. These datasets are derived from  the Caltrans Performance Measurement System
(PeMS) in California, which utilizes a variety of sensors, including inductive loops and radar, to capture precise traffic flow, speed, and occupancy information on highways every 30 seconds. These data are then integrated into 5 min.The details of the four datasets is shown in Table \ref{tb:1}.

\subsection{Experimental settings}
The FasterSTS we proposed is implemented on a dual-card NVIDIA RTX 2080Ti GPU using PyTorch 2.0.0. For the PEMS03, PEMS04, PEMS07 and PEMS08 datasets, we divide them into training, validation, and test sets in the same proportion 6:2:2 as in previous studies \cite{19,18}. Predictive samples are generated using a sliding windows, and future traffic flow for the next 12 time steps (60 minutes) is predicted using past traffic flow data from the previous 12 time steps (60 minutes), while other parameters are listed in Table \ref{tb:2}.
\begin{table}[htbp]
\tabcolsep=0.2cm
\renewcommand\arraystretch{1.2}%row height
\centering
\caption{Hyperparameters of FasterSTS model}
\begin{tabular}{ccccccccc}
\hline
Dataset & STSGCL & Batch size &H& D & learn rate\\ \hline
PEMS03 & 4 & 16& 32& 8& 0.001\\
PEMS04 & 6 & 16& 32 & 8 & 0.001\\
 PEMS07 & 2 & 16& 32& 10 & 0.001\\
 PEMS08 & 4 & 16& 32& 6 & 0.001\\  \hline
\end{tabular}\label{tb:2}
\end{table}

In order to evaluate the robustness and universality of the model more accurately, this paper uses Mean Absolute Percentage Error (MAPE), Mean Absolute Error (MAE), and Root Mean Square Error (RMSE) as the evaluation indexes. In general, the evaluation indexes are calculated with the following Eq.(\ref{eq:12}) to Eq.(\ref{eq:14}):\\
\begin{equation}\label{eq:12}
    MAPE \ = \ \frac{1}{n}\ \sum_i \left|{\frac{P_i \ - \ \hat P_i}{{P_i}}}\right| \times 100\%,
\end{equation}
\begin{equation}\label{eq:13}
  MAE = \frac{1}{n} \sum_i |{P_i} - {\hat P_i}|,\  
\end{equation}
\begin{equation}\label{eq:14}
    RMSE =  \sqrt{\frac{1}{n} \sum_i ({P_i} - {\hat P_i})^2},
\end{equation}
Where $P_i$ and $\hat P_i$ represent observed traffic flow and predicted traffic flow at node $n$ respectively.
\begin{table*}[htbp]
\centering
\renewcommand\arraystretch{1.5}
\tabcolsep=0.1cm
\caption{Evaluation results of each model on the datasets PEMS03 and PEMS04}
\begin{tabular}{cccccccccccccccc}     \hline
\multirow{2}*{Dataset}   & \multicolumn{3}{c}{\multirow{2}*{Model}}  & \multicolumn{3}{c}{15min} & \multicolumn{3}{c}{30min} & \multicolumn{3}{c}{60min}  & \multicolumn{3}{c}{Average} \\ \cline{5-16}
                        & \multicolumn{3}{c}{}                       & MAE    &MAPE(\%)   & RMSE        & MAE     &MAPE(\%)    & RMSE        & MAE     &MAPE(\%)    & RMSE        & MAE      &MAPE(\%)   & RMSE                \\ \hline
\multirow{9}*{PEMS03} 
                        & \multicolumn{3}{c}{VAR \cite{49}}                   &    17.41  &  18.20   & 25.42        & 22.13     & 24.28    & 32.20        &  31.65  & 37.42& 44.89        &     22.91  & 25.53    & 33.04          \\
                        & \multicolumn{3}{c}{LSTM \cite{12-1}}
                        & 16.69  &  16.02 & 25.54 & 20.03  & 19.34 & 30.40 & 27.42  & 29.48 & 40.20 & 20.64  & 20.57 & 31.63 \\
                        & \multicolumn{3}{c}{STGCN \cite{19}}
                        & 18.46 & 19.56&31.27& 19.55 &22.74& 32.59& 21.70  &24.08 &35.29&19.64  &21.63&32.74   \\
                        & \multicolumn{3}{c}{GWN \cite{18}}
                        & 16.06  & 15.39 & 27.38& 19.25  & 18.45 & 32.29& 26.12  & 26.32 &  41.89& 19.82  & 19.21 & 32.88\\
                        & \multicolumn{3}{c}{ASTGCN \cite{42}}
                        & 15.24  &15.31 &25.25&16.71 &15.64  &27.64&20.25  &18.63 &33.03&16.96  &15.97 &28.15  \\
                           & \multicolumn{3}{c}{STSGCN \cite{34}}
                         &15.98   & 16.34& 26.45& 18.06  & 17.96 &29.81& 22.15  & 21.77& 36.00 & 18.29  & 18.33 &30.07\\
                           & \multicolumn{3}{c}{STFGNN \cite{35}}               &      14.73 &  15.06   &  24.95        &  16.65 
 & 16.55 
     & 28.12 
         & 20.41 
   & 20.06 
     & 33.96 
       &
 16.54 & 17.15    & 26.25   \\
 & \multicolumn{3}{c}{STGODE \cite{47}}               &  13.96    &  15.43   & 25.93         & 16.35 & 16.32    & 27.38         & 19.54   & 18.79     & 30.97       &
16.39 & 16.54    &27.70   \\
     & \multicolumn{3}{c}{Bi-STAT \cite{46}}              & 14.30  & 14.78 & 25.07 & 15.43  & 15.67 & 2                                                                                  6.93  & 17.19 & 17.21 & 29.57  & 15.47 & 15.74 & 26.92\\
& \multicolumn{3}{c}{STWave \cite{ref-30}}              &13.91  & 14.90 & 24.82 & 14.92  & 15.53 & 26.70  & 16.68 & 18.92 & 29.19  & 14.96 & 16.54 & 26.62\\
     & \multicolumn{3}{c}{FasterSTS}              & 13.50  & 13.93  & 22.68 & 14.58  & 14.67 & 24.37  & 16.30 & 16.52  & 27.03  & 14.58 & 14.93  & 24.12\\ \hline

\multirow{9}*{PEMS04} 
                        & \multicolumn{3}{c}{VAR \cite{49}}
                         &22.75  & 16.95 & 33.11& 27.93   & 22.10 & 39.68& 38.26   &32.74 & 52.57& 28.78  & 23.01 & 40.72         \\
                        & \multicolumn{3}{c}{LSTM \cite{12-1}}
                        & 21.89  & 14.79 & 33.01& 25.98  &18.07 & 38.65& 35.12  & 27.28 & 50.35& 26.81 & 19.28 & 40.18\\
                        & \multicolumn{3}{c}{STGCN \cite{19}}
                        &23.43  &20.43 &35.30&25.45  &22.56 &37.83&30.45  &27.84 &44.12&25.91  &22.90&38.44\\
                        & \multicolumn{3}{c}{GWN \cite{18}}
                         &20.90  &14.44 &33.04&24.53  &17.45 &38.22&32.58  &23.38 &49.15&25.25  &18.22 &39.10\\
                        & \multicolumn{3}{c}{ASTGCN \cite{42}}
                        &19.55  &12.99 &31.16&20.32  &13.39 &32.39&23.63  &15.31 &37.21&20.59  &13.53&32.85\\
                            & \multicolumn{3}{c}{STSGCN \cite{34}}
                         & 19.95 &13.54 &32.01&21.48  &14.42 &34.28&24.54  &16.37 &38.59 &21.65  &14.54&34.49\\
                           & \multicolumn{3}{c}{STFGNN \cite{35}}               &  19.08    & 13.21  & 30.66        &  20.12 & 13.80     & 32.31        & 21.84   &  14.95     &  34.81       &
 20.11 &  13.78    & 32.19  \\
& \multicolumn{3}{c}{STGODE \cite{47}}               &  19.44    & 13.40  & 30.35         &  20.97 & 14.45     & 32.53        & 24.25   &  17.20     &  36.86      &
21.19&  14.65    & 32.85  \\
 & \multicolumn{3}{c}{Bi-STAT \cite{46}}               &  18.16    & 12.39   & 29.29         &  19.01 & 12.92     & 30.61        & 21.68  & 15.36    & 33.85      &
 19.66 & 13.02    & 30.64   \\
   & \multicolumn{3}{c}{STWave \cite{ref-30}}               &  17.64    & 11.89   & 28.98         &  18.68 & 12.62   & 30.62        & 20.03  &  13.68    & 32.64    & 18.64 & 12.62    & 30.55   \\
 & \multicolumn{3}{c}{FasterSTS }               &  17.70    & 11.75   & 28.73    &  18.55  & 12.27   &   30.04   & 19.64  &  12.98   &  31.67     &
 18.49 &   12.21  &  29.92    \\ \hline

\end{tabular}\label{tb:3}
\end{table*}

%%%%%%%%%%%%%%%%%%%%%%%%%%%%%%%%%%%%%%%%%%%%%%%%%%%%%%%%%%%%%%%%%%%%%%%%%%%%%%%%%%%%%%%%%%%%%
\begin{table*}[htbp]
\centering
\renewcommand\arraystretch{1.5}
\tabcolsep=0.1cm
\caption{Evaluation results of each model on the datasets PEMS07 and PEMS08}
\begin{tabular}{cccccccccccccccc}     \hline
\multirow{2}*{Dataset}   & \multicolumn{3}{c}{\multirow{2}*{Model}}  & \multicolumn{3}{c}{15min} & \multicolumn{3}{c}{30min} & \multicolumn{3}{c}{60min}  & \multicolumn{3}{c}{Average} \\ \cline{5-16}
                        & \multicolumn{3}{c}{}                       & MAE    &MAPE(\%)   & RMSE        & MAE     &MAPE(\%)    & RMSE        & MAE     &MAPE(\%)    & RMSE        & MAE      &MAPE(\%)   & RMSE                \\ \hline

\multirow{9}*{PEMS07} 
                        & \multicolumn{3}{c}{VAR \cite{49}}
                        & 25.10  &10.86 &37.22&31.30  &14.08 &45.23&42.96  &42.96 &60.04&32.03  &14.61 &46.12\\
                        & \multicolumn{3}{c}{LSTM \cite{12-1}}
                         &23.66  &10.48 &35.77&28.50  &12.76 &42.67 &39.12  &18.66 &56.60&29.32  &13.30&44.39\\
                        & \multicolumn{3}{c}{STGCN \cite{19}}
                         &32.52  & 16.23 &51.71&33.26  &16.25 &52.75&36.02  &17.59 &55.64& 33.62  &16.53 &53.04\\
                        & \multicolumn{3}{c}{GWN \cite{18}}
                         &21.72  &9.52 &34.78&25.89  &11.89 &41.05&34.83  &16.78 &53.72&26.58  &12.28 &41.85\\
                        & \multicolumn{3}{c}{ASTGCN \cite{42}}
                         &23.94  &10.25 &36.88&28.77  &12.50 &43.89&39.37  &18.00 &58.07&29.59  &13.04 &45.52\\
                            & \multicolumn{3}{c}{STSGCN \cite{34}}
                         & 22.65 &  9.73 & 36.40 & 25.20& 10.85  & 40.38 &30.57  & 13.40 &48.08&25.52	 &11.03&40.73 \\
                           & \multicolumn{3}{c}{STFGNN \cite{35}}               &  20.24 
    &  8.93 
  & 32.79 
        & 22.91 
 & 9.61 
     & 36.71 
       & 27.31 
   & 12.18 
    & 43.31 
      &
 22.80 
 & 10.03 
    & 36.69 
   \\
 & \multicolumn{3}{c}{STGODE \cite{47}}               &  21.29 
    &  9.63 
  & 34.62 
        & 23.69 
 & 10.74 
     & 38.41 
       & 28.73 
   & 13.27 
    & 45.73 
      &
 23.99 
 & 10.92 
    & 38.74 
   \\
      & \multicolumn{3}{c}{Bi-STAT \cite{46}}
                            &19.66  &8.36 &31.81&21.44  &9.07 &34.70&24.34  &10.43 &38.88&21.50  &9.15&34.64\\
    & \multicolumn{3}{c}{STWave \cite{ref-30}}
                            &18.23  & 7.61  &30.95&19.48  &8.16 &33.32&21.45  &9.07 &36.55&19.60  &8.21&33.29\\
   & \multicolumn{3}{c}{FasterSTS}
                            & 18.38  & 7.73  & 30.45 & 19.78  & 8.29  & 33.02 & 21.58   & 9.11 & 36.18 & 19.67   & 8.30 &  32.75 \\ \hline

\multirow{9}*{PEMS08} 
                        & \multicolumn{3}{c}{VAR \cite{49}}
                         &17.88  &12.17 &26.48&22.21&15.50  &32.39 &30.63  &21.94 &42.92&22.81&15.96 &32.97 \\
                        & \multicolumn{3}{c}{LSTM \cite{12-1}}
                        &17.71  &11.58 &26.65&21.31  &14.83 &31.97&29.37  &19.62 &42.53&21.96 &14.52 &33.22 \\
                        & \multicolumn{3}{c}{STGCN \cite{19}}
                        &20.53  &14.25 &29.92&22.21  &16.23 &32.24&26.40  &21.17 &37.37&22.62 &16.52 &32.66\\
                          & \multicolumn{3}{c}{GWN \cite{18}}
                          &15.59  &9.84 &25.27&18.28  &12.06 &30.02&24.18 & 15.63  &38.85&18.75  &12.12 &30.47\\
                        & \multicolumn{3}{c}{ASTGCN \cite{42}}
                        &16.36  &10.01 &25.24&18.38  & 11.12 &28.34&22.60  &13.57 &34.39&18.61  &11.27&28.82\\
                            & \multicolumn{3}{c}{STSGCN \cite{34}}
                          &16.60  &10.76 &25.37&17.75	  &11.56 &27.28&20.12  &13.04 &30.64&17.88  &11.71&27.30\\
                           & \multicolumn{3}{c}{STFGNN \cite{35}}               &   16.05   &  10.23   & 24.67       & 17.08 &  10.90     &  26.48       & 19.08   &  12.61   &   29.71        &
 16.33 &  11.12    &  26.50   \\
 & \multicolumn{3}{c}{STGODE \cite{47}}               &  15.48    &  9.86 & 23.68        & 16.84& 10.65     & 26.05       & 19.74   & 11.7    & 30.13      &
 17.01 & 10.68   & 26.25  \\
   & \multicolumn{3}{c}{Bi-STAT \cite{46}}               &  13.92    &  9.02   & 22.32         &  14.64 &  9.52     & 23.75        & 16.14  & 10.59     & 26.11       &
 14.76 & 9.62    & 23.80   \\
  & \multicolumn{3}{c}{STWave \cite{ref-30}}               &  12.77    &  8.50   & 21.71      &  13.69 &  9.40     & 23.47  & 14.98  & 10.44     & 25.85       &
 13.70 & 9.30    & 23.47   \\
  & \multicolumn{3}{c}{FasterSTS}               &  12.78    &  8.46   & 21.41         &  13.62  &  9.04     & 23.18        & 14.93  & 9.90  & 25.42       &
 13.60  & 9.01   &  23.10   \\ \hline

\end{tabular}\label{tb:4}
\end{table*}
\subsection{Baselines}
To validate the efficacy of the FasterSTS model in traffic flow prediction, this study selects eight state-of-the-art models widely employed in traffic flow forecasting:\\
$\bullet \textbf{Vector  \ Autoregression (VAR)}$\cite{49}: The VAR model is a time-series model that can handle the time-related correlation between multiple variables.\\
$\bullet \textbf{Long  \ Short \  Term \ Memory (LSTM)}$\cite{12-1}: The Long Short-Term Memory (LSTM) is a specialized recurrent neural network (RNN) architecture that effectively addresses the issue of long-term dependencies in RNNs.\\
$\bullet$ $\textbf{Spatio-Temporal}$ \ $\textbf{Graph}$ \ $\textbf{Convolutional}$ $\textbf{Convolutional}$ $\textbf{Networks(STGCN)}$\cite{19}:The STGCN model leverages temporal convolution blocks and spatial convolution blocks to enhance the extraction of both temporal and spatial features.\\
$\bullet \textbf{Graph \ Wavenet(GWN)}$\cite{18} :The proposed approach in this study, namely GWN, leverages adaptive static graphs to capture the inherent spatial correlations of road networks and integrates them with gated temporal convolutions for accurate predictive modeling of traffic flow.\\
$\bullet$ $\textbf{Attention}$ \ $\textbf{Spatio-Temporal}$ \ $\textbf{Graph}$ $\textbf{Convolutional}$ $\textbf{Network (ASTGCN)}$\cite{42}:Combines attention mechanisms with CNN and GCN to model both the dynamic spatial and temporal correlations inherent in traffic flow data.\\
$\bullet \textbf{Spatio-Temporal}$ $\textbf{synchronous}$ $\textbf{Graph}$ $\textbf{Convolutional}$ \ $\textbf{Network(STSGCN)}$\cite{34}:STSGCN can effectively capture local spatio-temporal correlations by utilizing the constructed spatio-temporal graph, as well as capture the heterogeneity of spatio-temporal data.\\
$\bullet \textbf{Spatio-Temporal}$ \ $\textbf{Fusion}$ \ $\textbf{Graph}$ \ $\textbf{Convolutional}$ \ $\textbf{Network (STFGNN)}$\cite{35}:STFGNN utilizes spatio-temporal fusion graph convolution and combines it with TCN to learn local and global spatio-temporal correlations separately.\\
$\bullet \textbf{Spatio-Temporal}$ \ $\textbf{Graph}$ \ $\textbf{ODE}$ \ $\textbf{Networks}$ \ $\textbf{for}$ $\textbf{Traffic}$ $\textbf{Flow}$ $\textbf{Forecasting(STGODE)}$  \cite{47}:Utilizes tensor-based ordinary differential equations to construct deep neural networks for simultaneous extraction of spatio-temporal features.\\
$\bullet  \ \textbf{Bidirectional} \ \textbf{Spatial-Temporal} \ \textbf{Adaptive} \ \textbf{Tr-} \\
          \textbf{ansformer} \ \textbf{{for}} \ \textbf{Urban}  \ \textbf{Traffic}   \ \textbf{Flow}  \ \textbf{Forecasting} \ (\textbf{Bi-} \\ \textbf{STAT})$ \cite{48} The encoder-decoder architecture integrates two major components: a temporal Transformer and a spatial Transformer. Its uniqueness lies in the addition of a recall module in the decoder, which aims to provide additional information support for the prediction task. Furthermore, we have designed the DHM module, which can flexibly adjust the complexity of the model according to the difficulty of the prediction task.
          $\bullet  \ \textbf{Decoupled}  \ \textbf{Traffic} \ \textbf{Prediction} \ \textbf{with}  \ \textbf{Efficient}   \\ \textbf{Spectrum-}  \textbf{based} \ \textbf{Attention} \ \textbf{Networks} \ (\textbf{STWave})$ \cite{ref-30} STWave decomposes complex traffic data into stable trends and fluctuating events, models them using dual-channel spatio-temporal networks, and predicts future traffic flow after fusion. Additionally, it incorporates a novel query sampling strategy and graph wavelet-based position encoding into the graph attention network to efficiently model dynamic spatial correlations.
\subsection{Experimental Results and Analysis}
Table \ref{tb:3} and \ref{tb:4} show the prediction results of FasterSTS compared with different baseline models on four datasets. The comparison results indicate that our proposed FasterSTS outperforms the state-of-the-art baseline models in terms of prediction performance and significantly leads all baseline models in three evaluation indexes on all four datasets.

Table \ref{tb:3} and \ref{tb:4} show the prediction results of all comparison methods on the PEMS03, PEMS04, PEMS07 and PEMS08 datasets at 30min, 45min, 60min and on average. From the tables, it can be seen that the prediction performance of the machine learning-based prediction methods HA and VAR is poor. LSTM has a significant improvement in prediction performance compared to the machine learning-based methods, but LSTM only treats traffic flow as time series and ignores the spatial correlation between different nodes, thus limiting its prediction performance greatly. STGCN and GWN use graph convolutional networks to capture the spatial correlation in the road network and combine the temporal feature extraction module to further capture the temporal correlation. Since they are spatio-temporal models, their performance is better than that of the temporal model LSTM. ASTGCN involves an attention mechanism into STGCN, and the attention mechanism can accurately model the dynamic spatial correlation. Therefore, its performance is better than that of STGCN and GWN. AGCRN introduces an adaptive graph into the spatio-temporal model, and AGCRN learns the hidden spatial correlation in traffic flow data in a data-driven manner. The adaptive graph can better fit the real road network, so its prediction performance is better than that of the spatio-temporal model based on a predefined graph,such as STGCN and GWN. 

STSGCN, STFGNN, and STGODE are spatio-temporal synchronous models that are designed based on prior knowledge in a predetermined manner to construct a spatio-temporal graph. They extract local spatio-temporal correlations by constructing spatio-temporal graph. Although they can capture the heterogeneity of spatio-temporal data, there are some shortcomings. First, STSGCN and STFGNN obtain a spatio-temporal graph by concatenating the temporal graph and the topological graph, which has a high time complexity and high computational cost. The temporal graph and the topological graph are generated based on the prior knowledge of domain experts in a predetermined manner, which contains some unreasonable descriptions, so the quality of the constructed spatio-temporal graph is poor. In addition, the spatio-temporal graph fusion rules in STSGCN and STFGNN are defined by researchers, which greatly limits the performance of spatio-temporal synchronous modeling. In contrast, STGODE constructs a spatio-temporal graph based on the temporal graph and the topological graph through a tensor-based ordinary differential equation. Although the quality of the spatio-temporal graph is still poor, the time complexity is decreased. However, they cannot directly extract long-term temporal correlations from historical data, but rather model long-term temporal correlations by gradually expanding the receptive field through stacking spatio-temporal layers. This results in further increase in time complexity. Finally, they are mainly limited to how to construct spatio-temporal synchronous modeling mechanisms, while ignoring the synchronous modeling of dynamic spatio-temporal correlations. 

The Bi-STAT model cleverly incorporates multi-head spatio-temporal attention mechanisms to precisely capture dynamic spatio-temporal correlations. Moreover, the model innovatively introduces a memory module, which recalls information from past moments to provide strong support for model prediction, thereby significantly enhancing the model's predictive performance. As a result, Bi-STAT has achieved a significant leap in model performance compared to the models mentioned above.The STWave model, similar to Bi-STAT, also adopts multi-head spatio-temporal attention to capture dynamic spatio-temporal correlations. However, STWave goes a step further by utilizing wavelet transform technology to skillfully decompose traffic flow into high-frequency and low-frequency components, and models these two parts separately to more finely depict the dynamic changes in traffic flow. This innovation has enabled STWave to achieve a significant improvement in performance compared to Bi-STAT.

FasterSTS can directly model local and long-term spatio-temporal correlations. Therefore, in comparisons with the baseline methods, it performs the best on the four datasets. FasterSTS improves the overall average prediction performance of STGODE by 10\%\~{}21\%, which is a very significant improvement in spatio-temporal synchronous modeling methods. Additionally, FasterSTS does not rely on fused spatio-temporal graph, so FasterSTS has a time complexity that is several times lower than that of STGODE and STFGNN.

To provide a more comprehensive visualization of the overall superiority of FasterSTS over the baseline model, we present a visualization of the performance of each model at each prediction steps in Fig.4. As shown in the figures, FasterSTS achieves the state-of-the-art prediction performance at all 12 prediction steps.

\subsection{Ablation  analysis}
In this section, we conducted ablation studies to verify the effectiveness of the fast graph operation, spatio-temporal synchronous convolution kernel, and the adaptive graph used in this paper. For this purpose, we set up five variants of FasterSTS. We conducted ablation experiments on the PEMS04 dataset, and the average performances of the four evaluation metrics are shown in Table \ref{tb:5}. These variants are as follows:

$\bullet$FasterSTS: This paper presents a spatio-temporal  synchronous modeling is approached.

$\bullet$w/o FGC: We discard the fast graph operations while keeping everything else unchanged in the FasterSTS.

$\bullet$w/o D: Remove the dynamic part of the spatio-temporal convolution kernel and replace it with a static adaptive embedding of the same size.

$\bullet$w/o DE: When building an adaptive graph, we do not build different adaptive graphs for each hidden dimension $D$.

$\bullet$w/o EP : Without using EP components, directly using $\Upsilon$ to capture temporal correlation and feature mapping, in this case $\Upsilon  \in {R^{T \times {d_{in}} \times d}}$.
\begin{table}[htbp]
\tabcolsep=0.2cm
\renewcommand\arraystretch{1.3}%row height
\centering
\caption{Ablation experiments on PEMS04.}
\begin{tabular}{cccccc}
\hline
PEMS04 & FasterSTS &w/o FGC &w/o D &w/o DE &w/o EP  \\ \hline
MAE & 18.49 & 18.40 & 18.65& 18.60& 18.55\\
RMSE & 29.92 &29.90 & 30.28 & 30.10& 30.05 \\
 MAPE & 12.21  &12.16  & 12.41 &  12.30& 12.26\\ \hline                   

\end{tabular}\label{tb:5}
\end{table}
%$\bullet$w/o DS: Replace the synchronous spatio-temporal graph convolution in FasterSTS with the traditional graph convolution.
%%%%%%%%%%%%%%%%%%%%%%%%%%%%%%%%%%%%%%%

%%%%%%%%%%%%%%%%%%%%%%%%%%%%%%%%%%%%%%%%%%
\begin{table*}[htbp]
\centering
\renewcommand\arraystretch{1.5}%row height
\tabcolsep=0.2cm%col width
\caption{Calculating cost on  PEMS07}
\begin{tabular}{cccccccccc}     \hline
\multirow{2}*{Dataset}   & \multicolumn{3}{c}{\multirow{2}*{model}}  & \multicolumn{3}{l}{Computation time} & \multicolumn{1}{l}{GPU Cost} \\ \cline{5-6} \cline{8-10}
                        & \multicolumn{3}{c}{}                       &Traning(s/epoch)    &Interferce(s)       &    & GPU memory(MB)      &       &     \\ \hline
\multirow{7}*{PEMS07(16)} 
& \multicolumn{3}{c}{STSGCN}                      & 665 & 115&  & 11287  \\
& \multicolumn{3}{c}{STFGNN}                      & 602  & 109 &  & 13555 \\
& \multicolumn{3}{c}{STGODE}                      & 256  & 34 &  & 6789 \\
                        & \multicolumn{3}{c}{Bi-STAT}                      & 740  & 134 &  & 41579 \\
                               & \multicolumn{3}{c}{STWave}                      & 452  & 78 &  & 9265 \\
                                                & \multicolumn{3}{c}{w/o FGC} & 51 & 10 &  & 2247 
                        \\
                        & \multicolumn{3}{c}{FasterSTS} & 37 & 9 &  & 1759 
                        \\\hline

\end{tabular}\label{tb:6}
\end{table*}

From the table, it can be seen that the predictive performance of FasterSTS is almost on par with that of w/o FGC, which fully demonstrates that the innovative fast graph operation in FasterSTS achieves a similar effect to traditional graph operations in aggregating information from other nodes. Even more notably, the fast graph operation significantly reduces time complexity and improves computational efficiency.
the predictive performance of FasterSTS is better than that of FasterSTS w/o D because dynamic embeddings are involved in the spatio-temporal synchronous graph convolution kernel, enabling the graph convolution process to capture dynamic temporal correlations simultaneously. The predictive performance of FasterSTS is better than that of FasterSTS w/o DE because FasterSTS builds different adaptive graphs with different hidden dimensions, and combines $\Psi$ with a likewise gated operation to dynamically adjust ${{\hat X}} = {\tilde D^{ - \frac{1}{2}}}\tilde A{\tilde D^{ - \frac{1}{2}}}{{X}}$, achieving the modeling of dynamic spatial correlations. The predictive performance of FasterSTS is better than that of FasterSTS w/o EP because the EP component in FasterSTS can model the implicit temporal correlation during the projection process.
%The predictive performance of FasterSTS w/o D is better than that of FasterSTS w/o DS because FasterSTS w/o D can extract temporal features in the graph convolution process.
\subsection{Hyperparameters analysis}
Fig.5 show the prediction performance of FasterSTS on different hyperparameters on the PEMS04 datasets. We only adjust one parameter of the model while keeping the other parameters as the best parameters of the model. $L$ represents the number of layers in the STSGCL. $H$ represents the hidden dimension of the model. $D$ represents the embedding dimension size of the adaptive graph. It can be seen that there is an optimal threshold for $L$, $H$, and $D$. When the values are smaller than the threshold, increasing the size of $L$, $H$, and $D$ can improve the prediction performance of the model. However, when the values are greater than the threshold, it leads to overfitting, and the model performance decreases. At the same time, the computational complexity of the model increases, and the computing efficiency decreases.
\subsection{Visualization}
Fig.6 illustrates a comparison between our model and the baseline model STGODE showcasing the prediction results over 500 steps on PEMS04 and PEMS08 datasets. It is evident that our model exhibits enhanced accuracy in predicting traffic peaks and valleys, particularly at their onset and conclusion. The FasterSTS combines dynamic spatio-temporal correlations modeling mechanism, which can effectively capture dynamic spatio-temporal correlations, enabling it to respond quickly to complex and changing traffic patterns.

\subsection{Computation Cost}
Let's briefly analyze the advantages of FasterSTS in terms of time complexity. By leveraging fast graph operations, we have reduced the time complexity of our proposed spatio-temporal synchronous graph convolution kernel from $Td_{in} \times Td_{out}$($Td_{in} \times Td_{out}$ is equal to $Td_{in}\frac{T}{2}d_{out} + Td_{out}\frac{T}{2}d_{in}$.) to $Td_{in}d_\Psi + Td_{out}d_\Psi$,$d_\Psi$ is much less than $\frac{T}{2}d_{out}$ and $\frac{T}{2}d_{in}$. The time complexity of adaptive graph operations has been decreased from $N^2$ to $2nN$, $2n$ is much less than $N$. Overall, the time complexity of the original model has been reduced from quadratic to linear level. To more intuitively demonstrate the advantages of FasterSTS in terms of time complexity and resource consumption, we continue to use the notation w/o FGC for the FasterSTS variant that discards fast graph computations. We compared the training and inference efficiency, as well as GPU memory consumption, of FasterSTS, w/o FGC, STGODE, STFGNN, and STSGCN on the PEMS07 dataset. To ensure a fair evaluation, all methods were run on a server equipped with an Intel Core i7 13700KF processor and a single NVIDIA RTX 2080 TI graphics card, with consistent batch sizes. The comparison results are shown in Table \ref{tb:6}. In terms of computational efficiency and resource utilization, FasterSTS outperforms the other compared methods by several times.

\section{Conclusion}
In this paper, we propose an efficient spatio-temporal synchronous model, FasterSTS. Firstly, we revolutionize the field by introducing a fast graph operation method that reduces the time complexity from the traditional quadratic level of graph operations to a linear level. Secondly, through meticulous design of the spatio-temporal synchronous graph convolution kernel, we are able to capture both static and dynamic temporal correlations while modeling spatial correlations, effectively capturing the spatio-temporal heterogeneity in the data. Then, based on the fast graph operation, we construct adaptive graphs using global and local adaptive embeddings to capture the static spatial correlations in the road network. Additionally, we cleverly design a form similar to a gating mechanism based on the spatio-temporal synchronous graph convolution kernel to dynamically adjust the extracted static spatial features, thereby modeling the dynamic spatial correlations in the road network. Finally, we conduct experimental comparisons on four real-world datasets, and the results demonstrate that our proposed FasterSTS significantly outperforms the state-of-the-art spatio-temporal synchronous models in terms of prediction performance, computational complexity, and resource consumption.
\ifCLASSOPTIONcaptionsoff
  \newpage
\fi

% trigger a \newpage just before the given reference
% number - used to balance the columns on the last page
% adjust value as needed - may need to be readjusted if
% the document is modified later
%\IEEEtriggeratref{8}
% The "triggered" command can be changed if desired:
%\IEEEtriggercmd{\enlargethispage{-5in}}

% references section

% can use a bibliography generated by BibTeX as a .bbl file
% BibTeX documentation can be easily obtained at:
% http://mirror.ctan.org/biblio/bibtex/contrib/doc/
% The IEEEtran BibTeX style support page is at:
% http://www.michaelshell.org/tex/ieeetran/bibtex/
%\bibliographystyle{IEEEtran}
% argument is your BibTeX string definitions and bibliography database(s)
%\bibliography{IEEEabrv,../bib/paper}
%
% <OR> manually copy in the resultant .bbl file
% set second argument of \begin to the number of references
% (used to reserve space for the reference number labels box)
\bibliographystyle{IEEEtran}      %IEEEtran为给定模板格式定义文件名
\bibliography{reference} %ref为.bib文件名
\vspace{-75 mm}
\begin{IEEEbiography}
[{\includegraphics[width=1in,height=1.25in,clip,keepaspectratio]{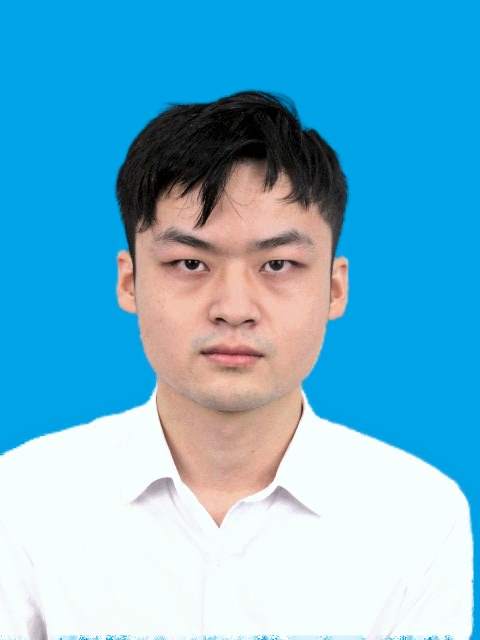}}]{Ben-Ao Dai} received the M.S. degree in Electronic Information at Zhejiang Sci-tech University, Hangzhou, Zhejiang, China,in 2024. He is currently working
toward the PhD degree in the Intelligent Transportation Systems Research Center, Wuhan University of Technology. His research interests include spatio-temporal data prediction,intelligent transportation system.
\vspace{-75 mm}
\end{IEEEbiography}
\begin{IEEEbiography}
[{\includegraphics[width=1in,height=1.25in,clip,keepaspectratio]{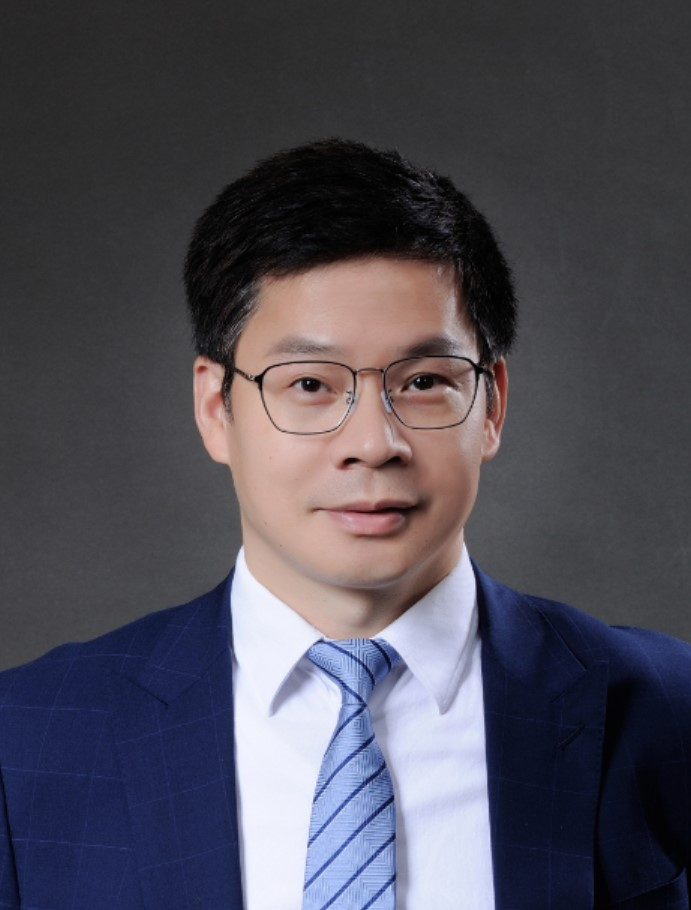}}]{Nengchao Lyu} is a professor of Intelligent Transportation Systems Research Center, Wuhan University of Technology, China. He visited the University of Wisconsin-Madison as a visiting scholar in 2008. His research interests include advanced driver assistance system (ADAS) and intelligent vehicle (IV), traffic safety operation management, and traffic safety evaluation. He has hosted 4 National Nature Science Funds related to driving behavior and traffic safety; he has finished several basic research projects sponsored by the National Science and Technology Support Plan, Ministry of Transportation, etc. He has practical experience in safety evaluation, hosted over 10 highway safety evaluation projects. During his research career, he published more than 80 papers. He has won 4 technical invention awards of Hubei Province, Chinese Intelligent Transportation Association and Chinese Artificial Intelligence Institute.
\end{IEEEbiography}

\vspace{-75 mm}
\begin{IEEEbiography}
[{\includegraphics[width=1in,height=1.25in,clip,keepaspectratio]{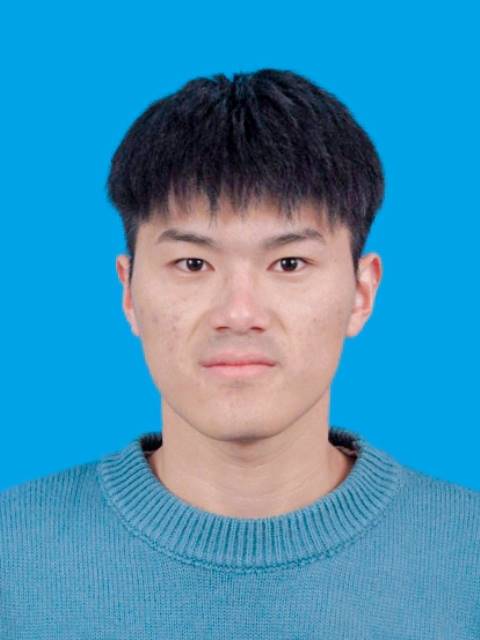}}]{Yongchao Miao} is currently pursuing the M.S. degree in Electronic Information at Zhejiang Sci-tech University,  Hangzhou, Zhejiang, China.  During his study for a 
master's degree.His research interest is intelligent transportation.
\vspace{-75 mm}
\end{IEEEbiography}
%
%
% biography section
% 
% If you have an EPS/PDF photo (graphicx package needed) extra braces are
% needed around the contents of the optional argument to biography to prevent
% the LaTeX parser from getting confused when it sees the complicated
% \includegraphics command within an optional argument. (You could create
% your own custom macro containing the \includegraphics command to make things
% simpler here.)
%\begin{IEEEbiography}[{\includegraphics[width=1in,height=1.25in,clip,keepaspectratio]{mshell}}]{Michael Shell}
% or if you just want to reserve a space for a photo:

%\begin{IEEEbiography}{Michael Shell}
%Biography text here.
%\end{IEEEbiography}

% if you will not have a photo at all:
%\begin{IEEEbiographynophoto}{John Doe}
%Biography text here.
%\end{IEEEbiographynophoto}

% insert where needed to balance the two columns on the last page with
% biographies
%\newpage

%\begin{IEEEbiographynophoto}{Jane Doe}
%Biography text here.
%\end{IEEEbiographynophoto}

% You can push biographies down or up by placing
% a \vfill before or after them. The appropriate
% use of \vfill depends on what kind of text is
% on the last page and whether or not the columns
% are being equalized.

%\vfill

% Can be used to pull up biographies so that the bottom of the last one
% is flush with the other column.
%\enlargethispage{-5in}

% that's all folks
\end{document}